\documentclass[ms]{agujournal2019}
\usepackage{url} 
\usepackage{lineno}
\usepackage{amsmath}
\usepackage{soul}
\usepackage{physics}
\usepackage{amssymb}
\usepackage{mathtools}
\usepackage{ mathrsfs }

\usepackage{xcolor}
\definecolor{gblue}{rgb}{0.0,0.7,0.6}

\definecolor{orange}{rgb}{1.0,0.5,0.0}

\journalname{Journal of Advances in Modeling Earth Systems}

\begin{document}

%
%


\title{4D-Var using Hessian approximation and backpropagation applied to automatically-differentiable numerical and machine learning models}

%
%




\authors{Kylen Solvik\affil{1}, Stephen G. Penny\affil{2}\affil{3}, Stephan Hoyer\affil{4}}


\affiliation{1}{University of Colorado Boulder, USA}
\affiliation{2}{Sofar Ocean, San Francisco CA, USA}
\affiliation{3}{Cooperative Institute for Research in Environmental Sciences at the University of Colorado Boulder}
\affiliation{4}{Google Research, Mountain View, CA 94043}




\correspondingauthor{Kylen Solvik}{Kylen.Solvik@colorado.edu }




\begin{keypoints}

\item The 4D-Variational data assimilation method is implemented using backpropagation of errors and Hessian approximation.
\item This low-cost approximate 4D-Var can be applied to any automatically differentiable forecast model and scales well to larger system sizes.
\item The new approach shows systematic advantages in both accuracy and computational speed over conventional 4D-Var in select scenarios. 

\end{keypoints}

%
%

%
%


\begin{abstract}
Constraining a numerical weather prediction (NWP) model with observations via 4D variational (4D-Var) data assimilation is often difficult to implement in practice due to the need to develop and maintain a software-based tangent linear model and adjoint model. One of the most common 4D-Var algorithms uses an incremental update procedure, which has been shown to be an approximation of the Gauss-Newton method. Here we demonstrate that when using a forecast model that supports automatic differentiation, an efficient and in some cases more accurate alternative approximation of the Gauss-Newton method can be applied by combining backpropagation of errors with Hessian approximation. This approach can be used with either a conventional numerical model implemented within a software framework that supports automatic differentiation, or a machine learning (ML) based surrogate model. We test the new approach on a variety of Lorenz-96 and quasi-geostrophic models. The results indicate potential for a deeper integration of modeling, data assimilation, and new technologies in a next-generation of operational forecast systems that leverage weather models designed to support automatic differentiation. 
\end{abstract}

\section*{Plain Language Summary}
There are two parallel communities advancing the state of the art in weather modeling - one developing conventional weather modeling and the other exploring machine learning (ML) models. Both communities have made exciting new advances. One is the development of automatically differentiable model implementations leveraging software tools like Julia and JAX. Another is the development of new ML weather models, which has generated excitement about the potential for using ML for weather prediction. However, today these ML models still rely on conventional numerical weather prediction (NWP) methods to produce their initial conditions, or `best estimate' of the current atmosphere from which the forecasts are initialized. Here, we use software tools designed for ML training algorithms to directly estimate those initial conditions from a combination of prior model forecasts and newly acquired observations - an approach that can be easily implemented with both conventional-style NWP models that newly support automatic differentiation as well as new ML weather models. We demonstrate that our approach performs well with two classic dynamical systems commonly used to test data assimilation methods.

%

%


%
%
%
%

\section{Introduction}

Data assimilation (DA) is a field that studies the optimal combination of theoretical knowledge of a physical dynamical system (such as the weather), usually encoded into a numerical model, with (typically noisy and sparse) observational data. One of the most common applications of DA is to provide initial conditions to numerical weather prediction (NWP) models used for operational forecasts. By combining past forecasts with newly available but sparse observations, a best guess (or `analysis') of the atmosphere can be determined that carries forward information from past observations and is more accurate and complete than either the forecast model or the observations alone. This process is cycled, and itself forms a dynamical system that is an important area of study in its own right \cite{carrassi2008, penny2017}.

While still in its infancy, the integration of artificial intelligence and machine learning (AI/ML) with DA is beginning to grow within the NWP community. A number of promising new ML weather models have been released, including NeuralGCM \cite{kochkov2023neural}, GraphCast \cite{lam2023}, PanguWeather \cite{Bi2023}, FourCastNet \cite{kurth2023}, and Spherical Fourier Neural Operators \cite{bonev_spherical_2023}, which all show potential for emulating global atmospheric weather dynamics and producing skillful forecasts. Based on simple metrics (typically a global RMSE score), these models already produce forecasts that rival the accuracy of the best physical-based NWP model forecasts \cite{rasp_weatherbench_2024}, and will presumably continue to improve until they offer full parity with operational NWP. 

All of these ML models currently require as initial conditions a full field input in the same form as the data they were trained on, which in most cases is a reanalysis product from the European Centre for Medium-Range Weather Forecasting (ECMWF). This limits their utility for making real-time operational forecasts. The question remains as to how these ML models can be used to make real-time operational forecasts, and whether there is a path to remove conventional NWP models as a dependency.

Here we demonstrate the utility of the JAX Python library \cite{jax2018github} in order to apply 4D-Var as a Gauss-Newton iteration. This is achieved by applying Hessian approximation using ML optimized gradient descent algorithms with backpropagation. We term this general class of algorithm ``Backprop-4DVar'', as it is agnostic to the specific choice of gradient descent method. This approach can be applied quite easily if the numerical model is written in JAX or any other software framework that supports automatic differentiation, such as PyTorch \cite{ansel_pytorch_2024} or Julia \cite{Julia-2017}. In some cases it may be quite difficult, or for all practical purposes impossible, to rewrite a model from scratch in a differentiable software language. Thus, we also provide an alternative approach using a trained differentiable ML surrogate model to take the place of the conventional numerical forecast model during the data assimilation optimization step.

\section{Methods}

\subsection{Data Models}

We test Backprop-4DVar using various implementations of two different chaotic dynamical systems: the Lorenz-96 system \cite{lorenz1996predictability} with dimensions ranging from 6 to 256, and two different implementations of 2-layer quasi-geostrophic (QG) baroclinic dynamics \cite{reinhold_dynamics_1982}. We consider the conventional numerical implementations of these dynamical systems as ``data generators'' to fit more closely to the standard paradigm of AI/ML methods training on preexisting datasets. The equations that define these dynamical systems are numerically integrated to produce the training, validation, transient, and test datasets. We specifically name the ``transient'' dataset to separate the data used for hyperparameter tuning (the validation set) and the data used for final evaluation of the methods (the test set), as (1) there may be correlations between immediately adjacent periods, and (2) in practice this is a more likely scenario for forecasting applications.

The Lorenz-96 system is a common testbed for evaluating DA algorithms for state estimation and forecasting of chaotic dynamical systems \cite{fertig2007, Abarbanel2010, Penny2014, penny2017, penny_integrating_2022, nerger2022, bach2023}. Its terms represent advection, diffusion, and forcing of a signal around a cyclic domain, usually conceptualized as a latitude ring around the globe. Our baseline configuration is a 36-dimensional (36D) Lorenz-96 using forcing $F=8.0$,
\begin{equation}
    \frac{\mathrm{d}u_k}{\mathrm{d}t} = u_{k-1}(u_{k+1} - u_{k-2})-u_k+F. \label{eq: lor96}
\end{equation}

We prepare the Lorenz-96 dataset by numerically integrating for 25,400 time steps with a $\Delta t$ of 0.01, which is 1,270 days in total based on the model time unit (MTU) defined by \citeA{lorenz1996predictability}. The first 14,400 time steps (two years) are used to spin-up the model toward the model attractor, starting from random initial conditions sampled from a Gaussian distribution with standard deviation 1.0. Because we do not train a machine learning forecast model for the Lorenz-96 case, we skip defining a training dataset. We generate an additional 11,000 time steps that are partitioned as validation (5,000 time steps), transient (1,000 time steps), and test (5,000 time steps) sets. Since the Tangent Linear Model (TLM) and adjoint of the Lorenz-96 are easily calculated analytically, we compare the performance of the conventional incremental 4D-Var and Backprop-4DVar on this dataset. 

To test Backprop-4DVar with a dataset produced by a more complex 2D spatially extended dynamical system, we use a quasi-geostrophic (QG) two-layer baroclinic model as implemented in PyQG-JAX \cite{otness_pyqg-jax_2024} (a JAX port of PyQG \cite{abernathay_pyqg_2022}) and the Python package `qgs' \cite{demaeyer_qgs_2020}. 

The 2-layer QG dynamics implemented in PyQG-JAX is driven by baroclinic instabilty of the base-state shear $U_1-U_2$,
\begin{align} \label{eq:pyqg_a}
\pdv{t}q_1 + \mathsf{J}\left(\psi_1, q_1\right) + \beta_1 \pdv{\psi_1}{x} + U_1\pdv{q_1}{x} = \hat{\delta}_s,
\end{align}
\begin{align} \label{eq:pyqg_b}
\pdv{t}q_2 + \mathsf{J}\left(\psi_2, q_2\right)+ \beta_2 \pdv{\psi_2}{x} + U_2 \pdv{q_2}{x} = -r_{ek}\nabla^2 \psi_2 + \hat{\delta}_s,
\end{align}
where $\hat{\delta}_s$ denotes small-scale dissipation, and the linear bottom drag in eqn. (\ref{eq:pyqg_b}) dissipates large-scale energy.
The potential vorticity is given by
\begin{align}
{q_1} = \nabla^2\psi_1 + F_1\left(\psi_2 - \psi_1\right), \\
{q_2} = \nabla^2\psi_2 + F_2\left(\psi_1 - \psi_2\right),
\end{align}
where
\begin{align}
F_1 \equiv \frac{k_d^2}{1 + \delta}, \qquad \text{and} \qquad F_2 \equiv \delta \,F_1.
\end{align}
The terms $\beta_1 = \beta + F_1\left(U_1 - U_2\right)$, and $ \beta_2 = \beta - F_2\left( U_1 - U_2\right)$ are the mean potential vorticity gradients, and the term
\begin{align}
k_d^2 \equiv\frac{f_0^2}{g'}\frac{H_1+H_2}{H_1 H_2}
\end{align}
is the deformation wavenumber, such that $H = H_1 + H_2$ is the total depth at rest.

The PyQG model is run with default parameters \cite{abernathay_pyqg_2022} with four different system dimensions: 512, 1152, 2048, and 8192 in gridded space (288, 624, 1088, and 2112 in spectral space, respectively). To spin-up the model, the PyQG model is run for 21,900 time steps---roughly 5 years with a time step of $\Delta t = 7200$ seconds---starting from random initial conditions based on \citeA{mcwilliams_emergence_1984}. The model is then run for an additional 9855 time steps (approximately 7.25 years in total) and then split into validation, transient, and test sets (1,095 / 4,380 / 4,380). The largest system size (8192D) is run with a shorter test period of 1,095 time steps to reduce computation time. 

The QG dynamics of the qgs package are based on \citeA{reinhold_dynamics_1982} and given as,
\begin{align}
    &\pdv{t}\overbrace{\mqty(\nabla^2 \psi_a^1)}^{\rm vorticity} + \overbrace{\mathsf{J}(\psi_a^1, \nabla^2 \psi_a^1)}^{\text{horizontal advection}} + \overbrace{\beta \pdv{\psi_a^1}{x}}^{\beta-\text{plane Coriolis force}} = -\overbrace{k'_d \nabla^2(\psi_a^1 - \psi_a^3)}^{\rm friction} + \overbrace{\frac{f_0}{\Delta p}\omega}^{\text{vertical stretching}} \\
    &\pdv{t}\mqty(\nabla^2 \psi_a^3) + \mathsf{J}(\psi_a^3, \nabla^2 \psi_a^3)+ \mathsf{J}(\psi_a^3, f_0 h/H_a) + \beta \pdv{\psi_a^3}{x} = k'_d \nabla^2(\psi_a^1 - \psi_a^3) -k_d \nabla^2 \psi_a^3 + \frac{f_0}{\Delta p}\omega
\end{align}
with the atmospheric streamfunctions $\psi_a^1/\psi_a^3$ at heights 250/750 hPa, and the vertical velocity $\omega = \dv{p}{t}$. Here, $\nabla = \pdv{x} \hat x+\pdv{y} \hat y$, $k'_d$ is the friction between layers, $k_d$ is the friction between the atmosphere and the surface, $h/H_a$ is the ratio of surface height to the characteristic depth of the atmosphere layer, $\Delta p = 500$ hPa is the pressure differential between the layers, and $\mathsf{J}$ is the horizontal Jacobian, defined as $\mathsf{J}(g_1, g_2) = \pdv{g_1}{x}\pdv{g_2}{y} - \pdv{g_1}{y}\pdv{g_2}{x}$.

To generate initial conditions, the qgs model is run for 200,000 time steps with a model time step of $\Delta t = 0.5$, starting from random initial conditions sampled from a uniform distribution between 0 and 0.001. We run an additional 121,000 time steps with (approximately 37 years in total) and then split into training, validation, transient, and test sets (100,000 / 10,000 / 1,000 / 10,000). The training dataset is used to train a machine learning surrogate model that is used in subsequent DA experiments.

\subsection{Forecast Models}

We use two different types of forecast model in the DA analysis/forecast cycle. First, we use numerical forecast models for the Lorenz-96 and QG dynamics, both implemented in JAX, and evolve the model dynamics using conventional numerical integration. For Lorenz-96 we use the JAX-based Dormand-Prince Runge-Kutta ordinary differential equation (ODE) numerical integration method \cite{shampine1986some, dormand1980family} with a time step $\Delta t = 0.01$. The TLM and adjoint of the Lorenz-96 model are determined analytically, coded explicitly, and calculated as a function of the nonlinear model trajectory. PyQG solves the two-layer QG system by applying a Fourier transform to the evolution equations and is integrated with a third-order Adams-Bashford scheme. For PyQG, the TLM and adjoint are computed explicitly using automatic differentiation. As an alternative, JAX allows us to directly use automatic differentiation with respect to the 4D-Var cost function rather than explicitly forming the TLM and adjoint, which we will describe in more detail in the next section.

In the second approach, we train a differentiable ML forecast model on a pre-existing training dataset describing QG baroclinic dynamics, generated using the qgs package. Because the qgs package does not currently support automatic differentiation, we instead use a reservoir computing (RC) architecture \cite{Jaeger01, Maass02, Jaeger02, Jaeger12, Luko12} as our ML surrogate model. Reservoir computing models are a simple type of recurrent neural network (RNN) architecture that have been demonstrated to excel at forecasting trajectories of multivariate dynamical systems \cite{Jaeger04, Vlachas20, platt_robust_2021,  penny_integrating_2022, platt_systematic_2022, platt2023}. Our reservoir computing model is implemented fully in JAX, enabling automatic differentiation.



The trained reservoir computing model is a self-contained dynamical system described by
\begin{equation}
    \mathbf{r}(t_n) = \alpha \tanh(\rho_a\mathbf{A} \mathbf{r}(t_{n-1}) + \sigma_u\mathbf{W}_{in} \mathbf{u}(t_{n-1}) + \mathbf{\sigma}_b) + (1-\alpha) \mathbf{r}(t_{n-1}).
\end{equation}
Here the reservoir state at time $t$ is given by $\mathbf{r}(t)$, while the inputs from the target dynamical system are given by $\mathbf{u}(t)$. The adjacency matrix $\mathbf{A}$ is specified randomly according to the sparsity macro-parameter (0.99 for our experiment), randomly setting $1 - \rho_s$ (where $\rho_s$ indicates the desired sparsity) of the entries to random values between $[-1, 1]$ and leaving all others as 0. The input matrix $\mathbf{W}_{in}$ is also generated randomly from a continuous uniform distribution between $[-1, 1]$ scaled by the macro-parameter $\sigma_u = 0.9877$. Both $\mathbf{A}$ and $\mathbf{W}_{in}$ are held fixed during training. The vector $\mathbf{\sigma}_b$ is a bias term that is trained. The remaining scalars are trained: $\alpha$ is a leak rate parameter, $\sigma_u$ is a gain term on the input matrix, and  $\rho_a$ is a gain term on the adjacency matrix used to rescale the matrix so that its spectral radius is equal to the trained spectral radius macro-parameter. These scalars are conventionally considered hyperparameters in older reservoir computing literature, however here they are more accurately described as macro-scale parameters because they are trained as part of the model optimization following best practices established by \citeA{penny_integrating_2022} and  \citeA{platt_systematic_2022}. We use pre-trained macro-scale parameter values that were trained for the quasi-geostrophic dynamics by \citeA{platt2023} using the covariance matrix adaption evolution strategy (CMA-ES) \cite{hansen2003, hansen23}.

The reservoir state is mapped to the state space of the target dynamical system via a readout operator $\mathbf{W}_{out}$ which maps $\mathbf{u}(t) = \mathbf{W}_{out}\mathbf{r}(t)$ for all $t$. Because we apply the pre-trained macro-parameters here with newly randomized adjacency and input matrices, it necessitates retraining $\mathbf{W}_{out}$. Because every element of $\mathbf{W}_{out}$ is trained, we call its elements the micro-scale model parameters. We retrain the micro-scale parameters on 100,000 time steps of the qgs model integration (the ``training'' set). The exact training parameters, as well as the pre-trained model weights, can be found in the Supporting Information.

Prior to commencing DA experiments, the reservoir computing system is spun up by providing accurate state information $\mathbf{u}(t_n)$ for $n \in [-N,0]$. For simplicity, we assume the system state is known up to time $t_0$. In practice the DA procedure itself would require a spinup time if started from scratch. After time $t_0$, the input state vector is replaced by $\mathbf{W}_{out}\mathbf{r}(t_{i-1})$, which when applied recursively produces a forecast.

\subsection{4D-Var Data Assimilation}

The conventional strong-constraint 4D-Var DA algorithm finds an optimal estimate of the initial state by minimizing the following cost function, consisting of background and observation terms:
\begin{equation}
    J(\mathbf{x}_0) = \frac{1}{2}(\mathbf{x}_0 - \mathbf{x}^b)^{T}\mathbf{B}_{0}^{-1}(\mathbf{x}_0 - \mathbf{x}^b) + \frac{1}{2}\sum_{i=0}^{N}(\mathbf{y}_{i}^{o} - H(\mathbf{x}_i))^T\mathbf{R}_{i}^{-1}(\mathbf{y}_{i}^{o} - H(\mathbf{x}_i)),
\end{equation}
where $x_b$ is the background estimate for $x_0$, the vector $\mathbf{y}_{i}^{o}$ contains the observations at time step $i$, and $H$ is the observation operator. 

The incremental form of the 4D-Var algorithm (Courtier et al. 1994) has been implemented operationally at ECMWF (e.g., Rabier et al. 2000). It is typically applied to solve the nonlinear least-squares problem as a multi-stage procedure using incremental corrections to the control variable and a series of linear least-squares approximations, which under certain assumptions is equivalent to the Gauss-Newton method \cite{lawless2005, lawless_inner-loop_2006, gratton_approximate_2007}. The outer loop integrates the nonlinear model forward in time to assess the fit to observations, and an inner loop uses a linearized version of the model (e.g. the tangent linear model) to perform a linear optimization, solving for $\delta{\mathbf{x}_0}$. The initial condition $\mathbf{x}_0$ is updated and then the outer loop is repeated $K$ times. 

Other Newton-type optimization methods have also been applied to 4D-Var, including Truncated Newton \cite{dimet_second-order_2002, wang_truncated_1995}, Inexact Newton \cite{dembo_inexact_1982}, Adjoint Newton \cite{wang_adjoint_1998}, and limited-memory quasi-Newton \cite{zou_numerical_1993}. Given poor initial conditions, the Gauss-Newton method of optimization can fail to converge. Line-search strategies and Gauss-Newton with regularization have been tested to alleviate this issue \cite{cartis_convergent_2021}. 

As mentioned above, the conventional strong-constraint incremental 4D-Var has an inner loop minimization step, which uses the TLM and adjoint to propagate perturbations within the optimization window. For this inner loop minimization we apply the JAX implementation of the BIConjugate Gradient STABilised (BI-CGSTAB) \cite{vanderVorst92} iterative solver to find the optimal correction to the initial condition based on the linearized dynamics. The steps are given below in detail:

\begin{itemize}
    \item  Integrate the numerical forecast model from the initial condition $\mathbf{x}(t_0) = \mathbf{x}_{0}^k$ (starting with $k=0$) to obtain the nonlinear forecast trajectory $\mathbf{x}(t)$, and compute the corresponding TLM and adjoint at regular intervals throughout the optimization window,

    \item  As the ``inner loop", minimize $J^k(\cdot)$ to obtain the incremental correction $\delta\mathbf{x}_{0}^{k}$ to the initial state:
    \begin{multline}
        J^k(\delta{\mathbf{x}_0^k}) = \frac{1}{2}(\delta{\mathbf{x}_0} - (\mathbf{x}^b - \mathbf{x}_0^{k}))^{T}\mathbf{B}_{0}^{-1}(\delta{\mathbf{x}_0} - (\mathbf{x}^b - \mathbf{x}_0^{k})) \\+ \frac{1}{2}\sum_{i=0}^{N}(H_i(\delta{\mathbf{x}_{i}^{k}}) - (\mathbf{y}_{i}^{o} - H_i(\mathbf{x}_{i}^k))^{T}\mathbf{R}_{i}^{-1}(H_i(\delta{\mathbf{x}_{i}^{k}}) - (\mathbf{y}_{i}^{o} - H_i(\mathbf{x}_{i}^k))
    \end{multline}
    In the inner loop, we optimise the cost function $J(\delta \mathbf{x})$ by setting the gradient $\nabla J_{\delta{\mathbf{x}_0}}$ equal to zero,
\begin{equation}
\nabla J^k(\delta{\mathbf{x}_0^k}) = \mathbf{B}^{-1}_0(\delta{\mathbf{x}_0^k}) -
\sum^{N}_{i=0}\mathbf{M}^\top_i\mathbf{H}^\top_i\mathbf{R}^{-1}_i(\mathbf{H}_i(\delta{\mathbf{x}_0^k}) - \mathbf{d}_i )^\top = 0
\label{equ:4Dvar-cf}
\end{equation}
where $\mathbf{M}$ is the TLM, $\mathbf{M}^\top$ is the adjoint, $\mathbf{H}_i$ is the linearization of $H_i$, and $\mathbf{d}_i=(\mathbf{y}_{i}^{o} - H_i(\mathbf{x}_{i}^k))$. We use the BI-CGSTAB linear solver, which solves equation (\ref{equ:4Dvar-cf}) after converting it to the form $\mathbf{Ax=b}$.

    \item Update the initial state $\mathbf{x}_0$:
    \begin{equation}
       \mathbf{x}_{0}^{k+1} = \mathbf{x}_{0}^k + \delta\mathbf{x}_{0}^k
    \end{equation}
    \item Repeat for $k = 0,...,K$ ``outer loops''
\end{itemize}

Theoretical stopping criteria for the inner loop have been defined by \citeA{lawless_inner-loop_2006}. In practice, the Gauss-Newton type optimization procedure is approximated (e.g., \citeNP{Lorenc_2000}; \citeNP{Mahfouf_2000}) by truncating the number of inner loop iterations, and by using an approximate linear system model (such as a reduced resolution). Such approximations are necessary to permit an operational implementation of 4D-Var to complete in the time limitations of the operational forecast-analysis cycle. Another approach that has been attempted is to approximate the linearized system by a low order inner loop minimization problem \cite{gratton_approximate_2007}.

\subsection{The Gauss-Newton method}

When the linearization of the forward model dynamics is exact, incremental 4D-Var is equivalent to the Gauss-Newton method  \cite{lawless_inner-loop_2006}. Following \citeA{gratton_approximate_2007}, Gauss-Newton minimizes a nonlinear functional $f(\mathbf{x})$,

\begin{equation}
\min_{\mathbf{x}} J(\mathbf{x}) = ||f(\mathbf{x})||_2^2.
\end{equation}

If we let $F(\mathbf{x}) = \frac{\partial f(\mathbf{x})}{\partial \mathbf{x}}$ be the Jacobian of $f(\mathbf{x})$, then the gradient $\grad J(\mathbf{x})$ and Hessian $\grad^2 J(\mathbf{x})$ are given by,

\begin{equation}
\grad J(\mathbf{x}) = F^T(\mathbf{x}) f(\mathbf{x})
\end{equation}

\begin{equation}
\grad^2 J(\mathbf{x}) = F^T(\mathbf{x}) F(\mathbf{x}) + G(\mathbf{x})
\end{equation}

where $G(\mathbf{x})$ represents higher order terms.

The stationary points of the nonlinear function $f(\mathbf{x})$ are found when 

\begin{equation}
\grad J(\mathbf{x}) =  F^T(\mathbf{x}) f(\mathbf{x}) = 0
\end{equation}

with the Hessian determining whether the critical points are local minima, maxima, or saddle points. For example, a critical point is a local minimum if the eigenvalues of the Hessian evaluated at that point are all real and positive. The Gauss-Newton method approximates the Newton method iteration by ignoring the higher order terms of the Hessian matrix, giving

\begin{equation}
\mathbf{x}_{k+1} = \mathbf{x}_k - (F^T(\mathbf{x}_k)F(\mathbf{x}_k))^{-1}F^T(\mathbf{x}_k)f(\mathbf{x}_k).
\end{equation}

The residual of this iteration is defined as,

\begin{equation}
\mathbf{r}_k = \mathbf{x}_{k+1} - \mathbf{x}_k = (F^T(\mathbf{x}_k)F(\mathbf{x}_k))^{-1}F^T(\mathbf{x}_k)f(\mathbf{x}_k).
\end{equation}

For high-dimensional systems the inverse of the approximate Hessian is difficult to compute directly. The Gauss-Newton algorithm is thus typically implemented as an iterative method aiming to drive the residual to zero,

\begin{equation}
F^T(\mathbf{x}_k)F(\mathbf{x}_k)\mathbf{r}_k =  -F^T(\mathbf{x}_k)f(\mathbf{x}_k),
\end{equation}

\begin{equation}
\mathbf{x}_{k+1} = \mathbf{x}_k + \mathbf{r}_k.
\end{equation}

The equivalence to the incremental 4D-Var minimization apparent if we set,

\begin{equation}
f(\mathbf{x}(t_0)) = \begin{pmatrix}
        \textbf{B}_0^{-1/2} (\mathbf{x}(t_0) - \mathbf{x}_b) \\
        \textbf{R}_0^{-1/2} (\mathbf{y}^o_0 - H_0(\mathbf{x}(t_0))) \\
               ... \\
        \mathbf{R}_n^{-1/2} (\mathbf{y}^o_n - H_n(\mathcal{M}_n(\mathbf{x}(t_0)))) \\
        \end{pmatrix}
\end{equation}

with the Hessian for 4D-Var consequently being,

\begin{equation}
F^T(\mathbf{x}(t_0))F(\mathbf{x}(t_0)) = \mathbf{B}^{-1}_0 + \mathbf{\hat{H}}^T\mathbf{\hat{R}}^{-1}\mathbf{\hat{H}}, \label{eq:4dvar_hessian}
\end{equation}

where

\begin{equation}
\mathbf{\hat{H}} = \left[ \left( \mathbf{H}_0 \right)^T, \left(\mathbf{H}_1 \mathbf{M}(\mathcal{M}_1(\mathbf{x}(t_0)),t_1,t_0) \right)^T, \ldots, \left(\mathbf{H}_n \mathbf{M}(\mathcal{M}_n(\mathbf{x}(t_0)),t_n,t_0) \right)^T
\right]^T,
\end{equation}

and

\begin{equation}
\mathbf{\hat{R}} = \begin{pmatrix} \mathbf{R}_1 & \mathbf{0} & \mathbf{0} \\
\mathbf{0} & \ddots & \mathbf{0} \\
\mathbf{0} & \mathbf{0} & \mathbf{R}_n \end{pmatrix}.
\end{equation}

The solution of this optimization finds the trajectory with the least squares fit to the observations, given the assumed known uncertainty and error characteristics of the background state and observations. In practice, as mentioned in the previous section, additional approximations are typically made for 4D-Var that use a truncated inner iteration linear solver, and the linearized least squares problem for determining the residual is replaced by a simplified or perturbed linear problem that can be solved more efficiently in the inner loop \cite{gratton_approximate_2007}.

\subsection{Automatic differentiation}

This work leverages the JAX software package designed for the Python programming language. JAX extends NumPy, which is the fundamental scientific computing package for Python. NumPy provides efficient computing by wrapping routines from compiled languages like C and Fortran with a Python interface. JAX extends NumPy by enabling computation on not just central processing units (CPUs), but also uses accelerated linear algebra (XLA) for just-in-time compilation to run efficiently on hardware accelerators  like graphics processing unit (GPUs) and tensor processing units (TPUs). 

We leverage a core feature of JAX, namely is its support for automatic differentiation of native Python code using an updated version of Autograd \cite{jax2018github}. JAX performs automatic differentiation (autodiff) by constructing a graph of all operations performed on the inputs and then differentiating each node of the graph. It supports native Python control flow operations like if statements and for loops. JAX also provides modules implementing NumPy and SciPy using JAX primitives, enabling further flexibility. In short, if a cost/loss function can be defined with a standard numerical library, then JAX can provide its gradient with respect to its control vector.

JAX extends the Autograd functionality to support both forward-mode and reverse-mode differentiation. JAX uses two foundational autodiff functions---one for Jacobian-Vector products (forward-mode) and another for Vector-Jacobian products (reverse-mode). JAX considers the Jacobian of a function $f(x)$ as a linear pushforward map that maps the tangent space at a point $x(t)$ to the tangent space at the point $f(x(t))$. Given a point $x$ and a tangent vector $v$, the Jacobian-vector product (JVP) is: $(x,v) \rightarrow \partial f(x) v$. Similarly, a pullback map can be formed for the vector-Jacobian product (VJP) $(x,v) \rightarrow v \partial f(x)$, which when considered as a matrix operation is equivalent to the adjoint conjugate  $(x,v) \rightarrow \partial f(x)^T v$. 

In the JAX software, the JVP/VJP formalism permits building the Jacobian matrices one row at a time, with the floating point operation cost for evaluating the adjoint to be only about three times the cost of evaluating the function $f(x)$. Backpropagation (a.k.a. reverse-mode differentiation) is achieved by applying a sequence of matrix-free VJPs. A drawback is that memory scales with the depth of the computation. However, for Backprop-4DVar we will limit the depth by restricting the optimization to be applied independently for each data assimilation cycle in sequence.

\subsection{Backprop-4DVar Data Assimilation}

As an alternative to the conventional 4D-Var two-stage multi-loop procedure, we propose a new approach inspired by ML optimization methods that leverages the JAX Autograd capability. We perform the 4D-Var optimization using an optimized gradient descent, Hessian approximation, and backpropagation of error, applied to the 4D-Var cost functional $J$ interpreted as a loss function, which we term ``Backprop-4DVar''. Rather than separately performing an outer loop (with the current best guess for $\mathbf{x}_0$) and an inner loop (using a linearized version of the model to optimize for $\delta{\mathbf{x}_0}$), Backprop-4DVar replaces the inner loop with the backpropagation of errors and explicit calculation of gradients of the loss function with respect with respect to the state vector at time $t_{0}$ ($\nabla {J(\mathbf{x}_0)}$). Using either a numerical model or reservoir computing ML model written in JAX, it is trivial to compute the necessary gradients using JAX's automatic differentiation capability. The algorithm is presented below. Compared with the conventional incremental 4D-Var algorithm, the implementation is incredibly simple, as much of the complexity has been encapsulated in the JAX backpropagation routines:

\begin{itemize}
    \item Integrate the numerical forecast model to obtain the nonlinear forecast trajectory $\mathbf{x}(t)$ and compute gradients of the cost function $J$ with respect to $\mathbf{x}_0$:
    \begin{equation}
       \nabla{J(\mathbf{x}_{0}^{k})} = \mathtt{jax.grad}(J)(\mathbf{x}_0^k) 
    \end{equation}
    \item Calculate an approximation of the Hessian for 4D-Var (see Eq. \ref{eq:4dvar_hessian} for full Hessian). Here we consider the approximation: 
    \begin{equation}
        F^T(\mathbf{x}(t_0))F(\mathbf{x}(t_0)) \approx \alpha^{-1} \left[\mathbf{B}^{-1} + \mathbf{H}_0^T\mathbf{R}^{-1}\mathbf{H}_0 \right]\label{eq:hessian_approx}
    \end{equation}

    \item Update $\mathbf{x}_0$:
    \begin{equation}
        \mathbf{x}_{0}^{k+1} = \mathbf{x}_{0}^k - \left[F^T(\mathbf{x}(t_0))F(\mathbf{x}(t_0)) \right]^{-1} \nabla{J(\mathbf{x}_{0}^{k})} \label{eq:backprop_step}
    \end{equation}
    \item Repeat for $k = 0,...,K$ iterations (i.e. gradient descent steps, analogous to `outer loops' in incremental 4D-Var)  
\end{itemize}

We note that if the Hessian were approximated as $\alpha^{-1}\mathbf{I}$ then equation \eqref{eq:backprop_step} would the standard gradient descent method. We also note that the full Hessian can be calculated automatically using JAX, though this is a costly operation. We test two versions of Backprop-4DVar: one with the approximate Hessian defined in equation \eqref{eq:hessian_approx} and one with the exact Hessian computed using JAX automatic differentiation. 

To tune the learning rate $\alpha$ and per-iteration exponential decay rate, we perform a Bayesian search on the validation datasets using hyperopt \cite{Bergstra_2015} and RayTune \cite{liaw2018tune}. This tuning is performed once and then the same $\alpha$ and decay rate is used throughout all cycles of the DA experiment for the test period. 

For all experiments, the search space for the learning rate $\alpha$ is $e^{-5}$ to $1.0$ and for the exponential learning rate decay factor is 0.1 to 0.99. For all experiments execept for the Lorenz-96 observation error experiments, he search is conducted with 50 samples used per experiment configuration (e.g. 50 samples each for the three system sizes in the PyQG-Jax experiments). For the Lorenz-96 observation error experiments, the RayTune search uses 20 samples per configuration (each combination of observations standard deviation and number of observations). The smaller number of samples is used to reduce execution time given the large number of experiment configurations. For the largest PyQG-JAX system (8192D), we do not perform tuning and use $\alpha = 0.5$ and $\alpha_{decay} = 0.5$, which were found to be adequate first-guesses for the smaller PyQG-JAX system dimensions.

The use of JAX's Autograd autodifferentiation bypasses the need for explicitly determining the TLM and adjoint of the forward model dynamics, allowing for the automatic calculation of gradients used for backpropogation of errors from the departure terms in the loss function to the initial conditions. As long as the observation operator $H(\cdot)$ is implemented in the JAX framework, its explicit linearization is also not required.

\subsection{Experiment Design}

We compare 4DVar, Backprop-4DVar with the exact Hessian, and Backprop-4DVar with the approximate Hessian over a range of experimental conditions. 
For all experiments, the background and observation error covariance matrices are defined as diagonal matrices with values of $\sigma_{background}^2$ and $\sigma_{observation}^2$ respectively.

We first evaluate the performance of each DA method on a 36-dimensional Lorenz-96 system while varying the degree of observation error and the sparsity of observations. We repeat the data assimilation experiments with 6, 12, 18, 24, 30, and 36 observations per cycle and observation noise standard deviation ranging from 0.1 - 2.0 model units. A total of 30 separate trials are run for each experiment configuration (i.e. each unique combination of observation number and noise), with 4D-Var and Backprop-4DVar given identical inputs for each trial. For each trial, the observation locations are randomly selected at the beginning of the experiment and are held fixed across all cycles. In addition, Gaussian noise with standard deviation of 1.0 is added to the initial conditions prior to commencing the data assimilation experiments. The analysis window is 10 time steps. In all cases, we set $\sigma_{observation}$ as 1.25 multiplied by the observation noise standard deviation as an overestimate, assuming in practice the observation error is unknown, and we set $\sigma_{background}$ as the observation noise standard deviation divided by 1.5, based on tuning experiments. Given the use of a diagonal error covariance matrices, only the ratio of background to observation error is relevant.

To further compare 4D-Var and Backprop-4DVar, we run data assimilation experiments on a broader range of Lorenz-96 system settings. First, to investigate how accuracy and computation time scale with system dimension, we repeat the experiment with increasing system dimensions: 6, 20, 36, 72, 144, and 256. In all cases, observations are sampled at half of the grid points (selected randomly at the beginning of the experiment and fixed across all cycles) every 5th time step and with added random Gaussian noise with standard deviation = 0.5. To obtain a more accurate comparison, we repeat each of these experiments 50 times and calculate the mean RMSE and computation time.

Next, to evaluate the performance of Backprop-4DVar on a more complex model, we perform data-assimilation experiments on the two-layer quasi-geostrophic system implemented in the PyQG JAX port with increasing system sizes with dimension: 512, 1152, 2048, and 8192 in gridded space (corresponding to 288, 624, 1088, and 2112 modes in spectral space, respectively). Using an analysis window of 6 time steps ($\Delta t = 2$ hours), we sample observations every 3 time steps, with Gaussian noise added using variable-specific standard deviation equal to 10\% of the per-variable climatological standard deviation over the spin-up period. In the background error covariance matrix, $\sigma_{background}$ is set to 0.05 times the per-variable standard deviations and $\sigma_{observation}$ is set as 0.125 times the per-variable standard deviation. For 4D-Var, the tangent linear model and adjoint model are calculated using the JAX reverse-mode automatic differentiation. Due to computational limitations, only Backprop-4DVar with the approximate Hessian is applied and evaluated on the highest resolution (8192D) QG system.

For the qgs quasi-geostrophic dynamics, the forecast phase of the DA cycle is performed using the trained RC model. Following \citeA{penny_integrating_2022}, the DA is performed in the reservoir space (or ``hidden space"). The initial state $\mathbf{x}_0$ and forecast trajectory $\mathbf{x}(t)$ are defined in the reservoir space. In order to compare with the observations sampled from the nature run in the system space, the observation operator $\mathbf{H} = \mathbf{S} \mathbf{W}_{out}$ maps the reservoir states into the system space, where $\mathbf{W}_{out}$ is the readout operator of the trained RC model and $\mathbf{S}$ is a matrix that maps from the full 20D system space to the locations of the observations. Each iteration updates $\mathbf{x}_0$ and computes the updated forecast trajectory $\mathbf{x}(t)$ in the reservoir space. For the DA experiment setup, observations are sampled from 10 of the 20 variables at every 3rd time step. Observational noise is added by first calculating the climatological standard deviation for each of the 20 variables throughout the training period and then adding random Gaussian noise with a standard deviation of 10\% of this per-variable climatological standard deviation to each observed variable. The analysis window is 9 time steps. $\sigma_{background}$ is set to the per-variable standard deviations in the reservoir space multiplied by 0.1 and $\sigma_{observation}$ is the per-variable standard deviations in the system space multiplied by 0.125. Generating the initial conditions is made more complex when using the RC model because the DA is performed in the hidden space. Following \citeA{penny_integrating_2022}, before using the RC model to make forecasts, it must first be spun up by synchronizing it with the QG dynamics. We use the transient period to do this synchronization, and apply Gaussian noise (SD = 0.001) to the transient nature run data in order to perturb the hidden space of the forecast model away from the true state at the starting time of the DA experiment. In practice, this spin up procedure could be applied using recent forecasts, with initial conditions having been constrained to observations via DA.

\section{Data}

 We demonstrate the aforementioned methods by using Lorenz-96 and QG dynamics. Implementations of these systems are numerically integrated to generate datasets that either serve as `truth' (called the `nature run') to be sampled in order to provide sparse and noisy observations, or as a training/validation/transient/testing dataset to be used for ML experiments.

 In addition to the train/validation/test breakdown, we add the concept of a `transient' period. Often, it is possible that a multivariate time series may have correlations with the neighboring periods. Thus, we suggest the breakdown of a dataset to be: train, validation, transient, test to provide additional confidence that the test period is truly independent from choices influenced by the validation dataset. The presence of this transient period also imitates the realistic scenario that any ML solution will have a delay between training and deployment. We acknowledge the possibility that an application could be designed to be trained online and then immediately make a prediction, but expect that this is not the simplest use case and should be stated explicitly.

\section{Results}

\subsection{Lorenz 96}

\begin{figure}
\noindent\includegraphics[width=\textwidth]{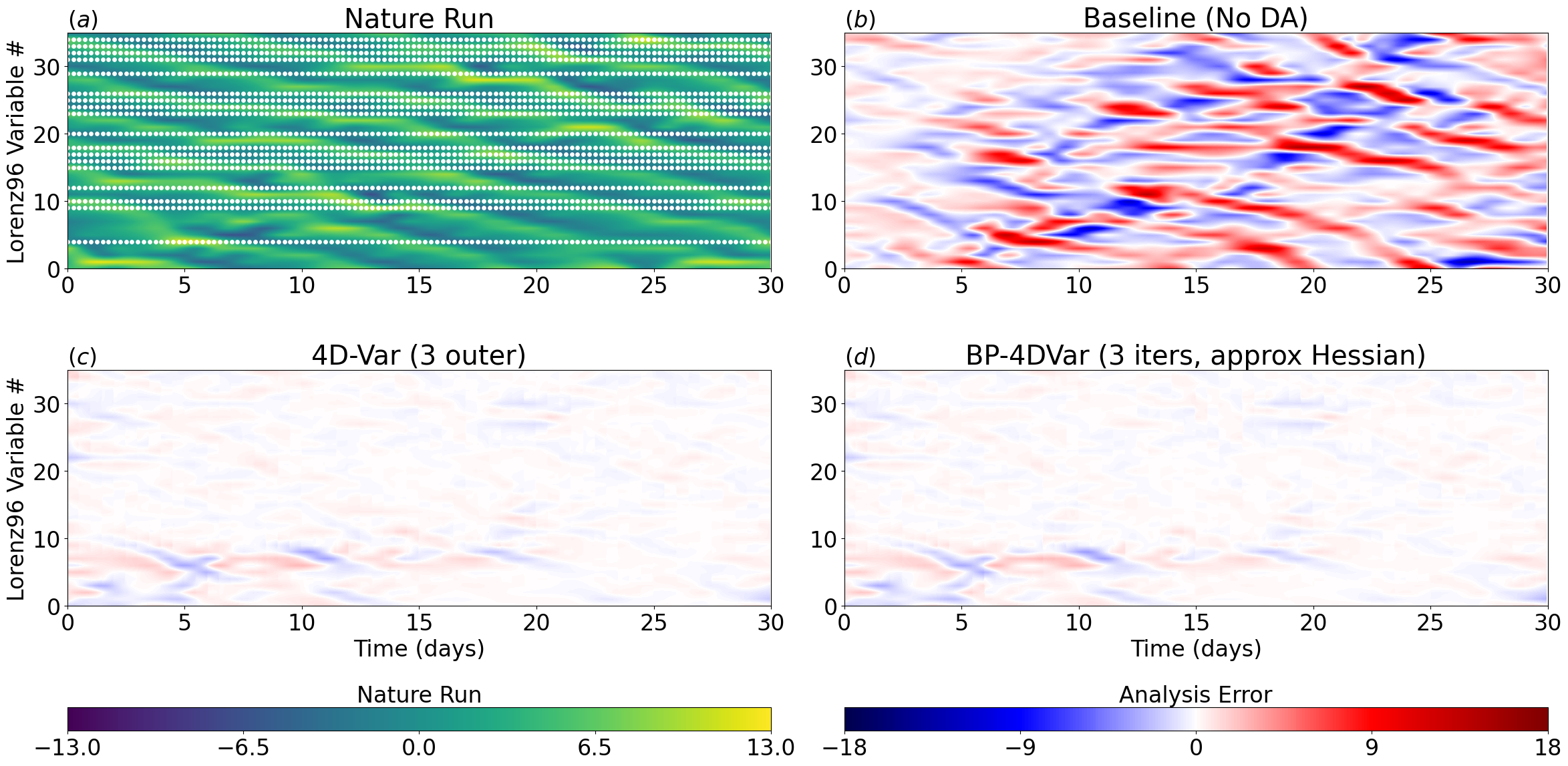}
\caption{(Top left) An example of the evolution of the 36-dimensional Lorenz-96 system over 600 time steps ($\Delta t=0.01$), or 30 days if one model time unit (MTU) corresponds to 5 days. The nature run is shown with observation locations at their appropriate times as white dots. (Top right) A baseline run without data assimilation shows the exponential error growth that occurs when the model is initialized from imperfect initial conditions. (Bottom left) the conventional incremental 4D-Var is used to assimilate the observations to reconstruct a state estimate that remains close to the nature run. (Bottom right) The Backprop-4DVar produces state estimates very close to the 4D-Var reconstruction.}
\label{l96_progression}
\end{figure}

We begin with an illustrative example showing the evolution of the Lorenz-96 system nature run (Figure \ref{l96_progression}, top left), compared with a forecast model initialized with imperfect initial conditions (Figure \ref{l96_progression}, top right). This example is taken from the start of the test dataset, after 20,400 total time steps of spin-up from randomized initial conditions. Because the system is chaotic, the errors grow rapidly until they saturate at a maximum that is determined by a range of states that the system can produce. The DA methods 4D-Var and Backprop-4DVar both use the `perfect' numerically integrated Lorenz-96 equations as the forecast model. The 4D-Var (Figure \ref{l96_progression}, bottom left) and Backprop-4DVar (Figure \ref{l96_progression}, bottom right) are applied using observations sampled at 18 of the 36 model grid points every 5 model time steps (with locations illustrated on the Nature Run panel). The analysis window is 10 model time steps. All runs start from the same perturbed initial conditions with Gaussian noise with standard deviation 1.0, and the same observations perturbed with Gaussian noise with standard deviation 0.5 (observations). In this case, we see qualitatively that 4D-Var and Backprop-4DVar have very similar performance and both have much lower RMSE than the control/baseline model run without data assimilation. 

\begin{figure}
\noindent\includegraphics[width=\textwidth]{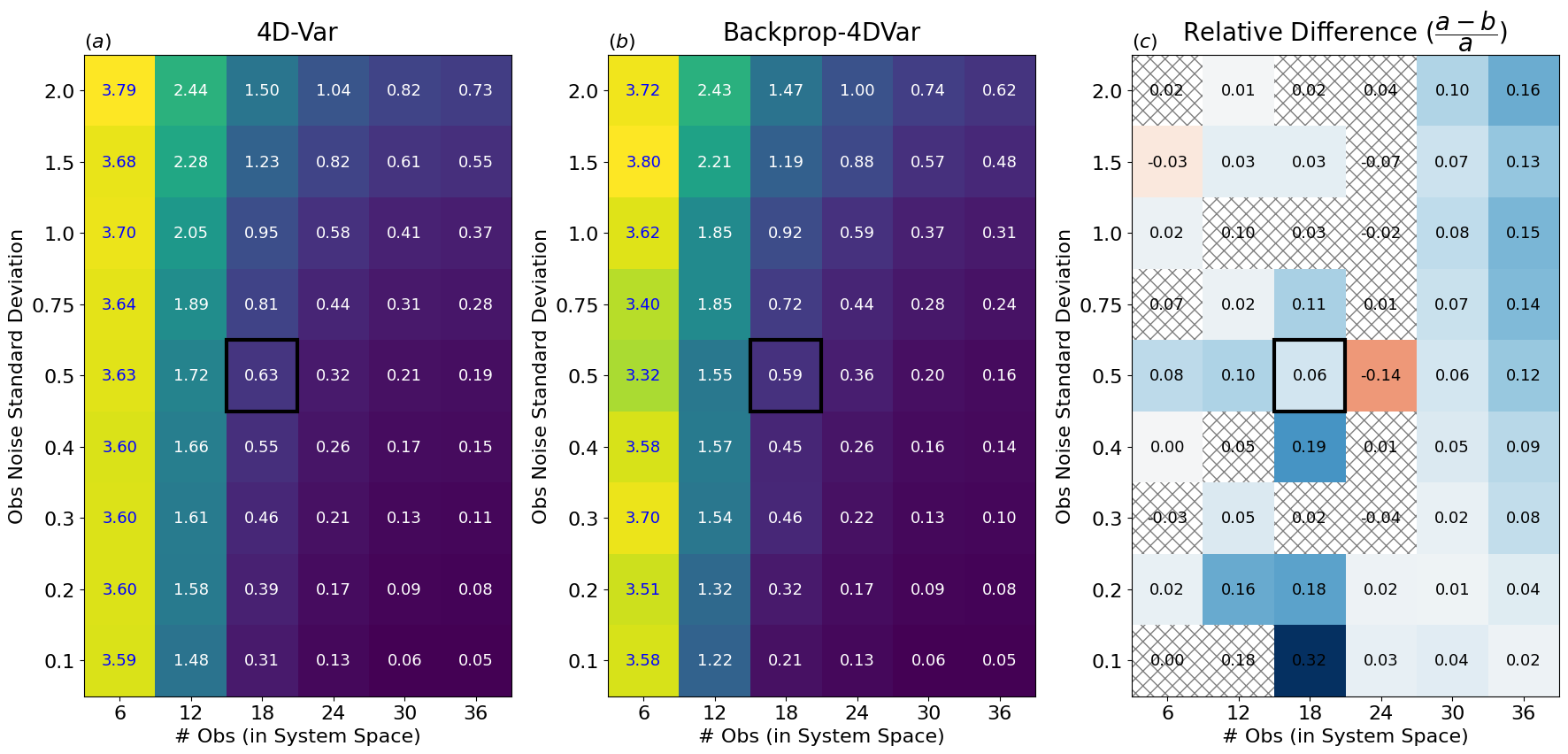}
\caption{A comparison of RMSE for the conventional incremental 4D-Var and Backprop-4DVar with varying observation coverage (x-axis) and observation noise (y-axis). A total of 1620 experiment results are summarized here, with 30 trials run for each combination of observation number and noise (i.e. each cell in the heatmap). The mean of the difference in RMSE, normalized by dividing by the 4D-Var RMSE, is shown in panel (c). standard deviation of differences in RMSE are shown in panels 3 and 4. Panel (c) shows the mean relative RMSE difference between the two, calculated by from dividing the absolute difference by the 4D-Var RMSE. Paired Student's t-tests are run for each cell using $\alpha = 0.01$. Cases where $p > \alpha$ and thus do not pass the conditions of the test are greyed out, while colored cells represent cases where $p < \alpha$. Positive values, shown in blue, indicate that on average Backprop-4DVar has a lower RMSE than 4D-Var. Negative values in red indicate that 4D-Var has a lower RMSE than Backprop-4DVar. The cell outlined in bold black has the same experimental configuration (36D with 18 observations and observation noise standard deviation of 0.5) as the 36D experiment shown in Figure \ref{l96_timecomp}}.
\label{l96_heatmap}
\end{figure}

We next expand this investigation to examine the impact of varying the observation coverage and observation noise (Figure \ref{l96_heatmap}), again using system size 36. Our objective is to verify that Backprop-4DVar produces results of a similar accuracy to the conventional 4D-Var. The experiment results are calculated over 30 trials for each entry. We note that the results from Figure \ref{l96_progression} represent a single case that corresponds to the cell with 18 observations and 0.5 observation noise standard deviation, roughly in the center of Figure \ref{l96_heatmap}, outlined with a bold black box. 

With greater observation coverage and lower observation noise, Backprop-4DVar and 4D-Var have very similar error. In cases with sparse observation coverage and lower observation noise, there is no consistent winner: in some cases 4DVar has lower error, in other cases Backprop-4DVar has lower error. In cases with more observations and more noise, Backprop-4DVar appears to outperform the standard incremental 4D-Var. With sparser observations, 4D-Var occasionally experiences transient filter divergence and this is reflected in larger cumulative RMSE statistics. The results show that Backprop-4DVar is generally as effective as the conventional incremental 4D-Var at identifying the optimal initial state in each DA cycle.

\begin{figure}
\noindent\includegraphics[width=\textwidth]{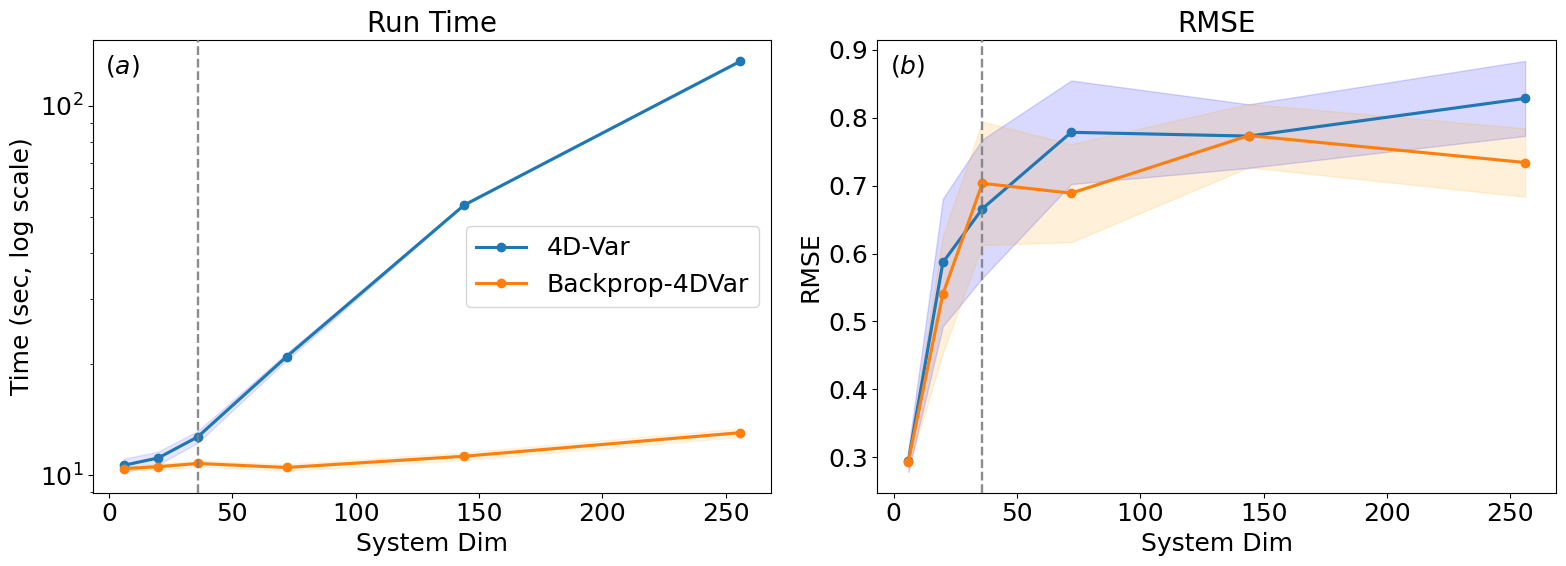}
\caption{(a) Comparison of 4D-Var and Backprop-4DVar runtimes, and (b) RMSE with increasing Lorenz 96 system size. The mean of results from 50 isolated trials with randomized initial conditions and observations are shown with $95\%$ confidence intervals shaded. For reference, the grey line highlights the experiment with the same configuration as the black-outlined cell in Figure \ref{l96_heatmap} (36D with 18 observations and observation noise standard deviation of 0.5)}
\label{l96_timecomp}
\end{figure}

As the system dimension increases, the RMSE of 4D-Var and Backprop-4DVar diverge somewhat but without one clearly outperforming the other. However, we see a substantial difference in run time between the conventional incremental 4D-Var and Backprop-4DVar (Figure \ref{l96_timecomp}). The conventional incremental 4D-Var relies on repeated matrix-vector multiplications, which scales quadratically in computation time with respect to system size. It is well known that in practice 4D-Var must be implemented with functional replacements for these costly operations. We are able to avoid such implementation modifications with Backprop-4DVar, as it scales more linearly with respect to system size. We do not make any attempts to optimize the conventional 4D-Var implementation here; an evaluation at the scale of an operational NWP forecast system is left for future work as it will likely depend on a number of additional factors, including the specific target hardware platform.

\subsection{Two-Layer Quasi-geostrophic model implemented in PyQG-Jax}

We next compare the performance of the DA methods applied to QG dynamics. Overall, the results are qualitatively the same. Both 4D-Var and Backprop-4DVar produce state estimates of the QG system with similar accuracy, while the Backprop-4DVar is considerably less costly.

In the following DA experiments, the analysis window is 6 time steps, while we sample observations every 3 time steps. Observations have Gaussian noise added using variable-specific standard deviation equal to 10\% of the per-variable climatological standard deviation, calculated over the spin-up period. The $\sigma_{background}$ error is set to 0.05 times, and $\sigma_{observation}$ to 0.125 times, the per-variable climatological standard deviations.

\begin{figure}
\begin{center}
\noindent\includegraphics[height=0.85\textheight,keepaspectratio]{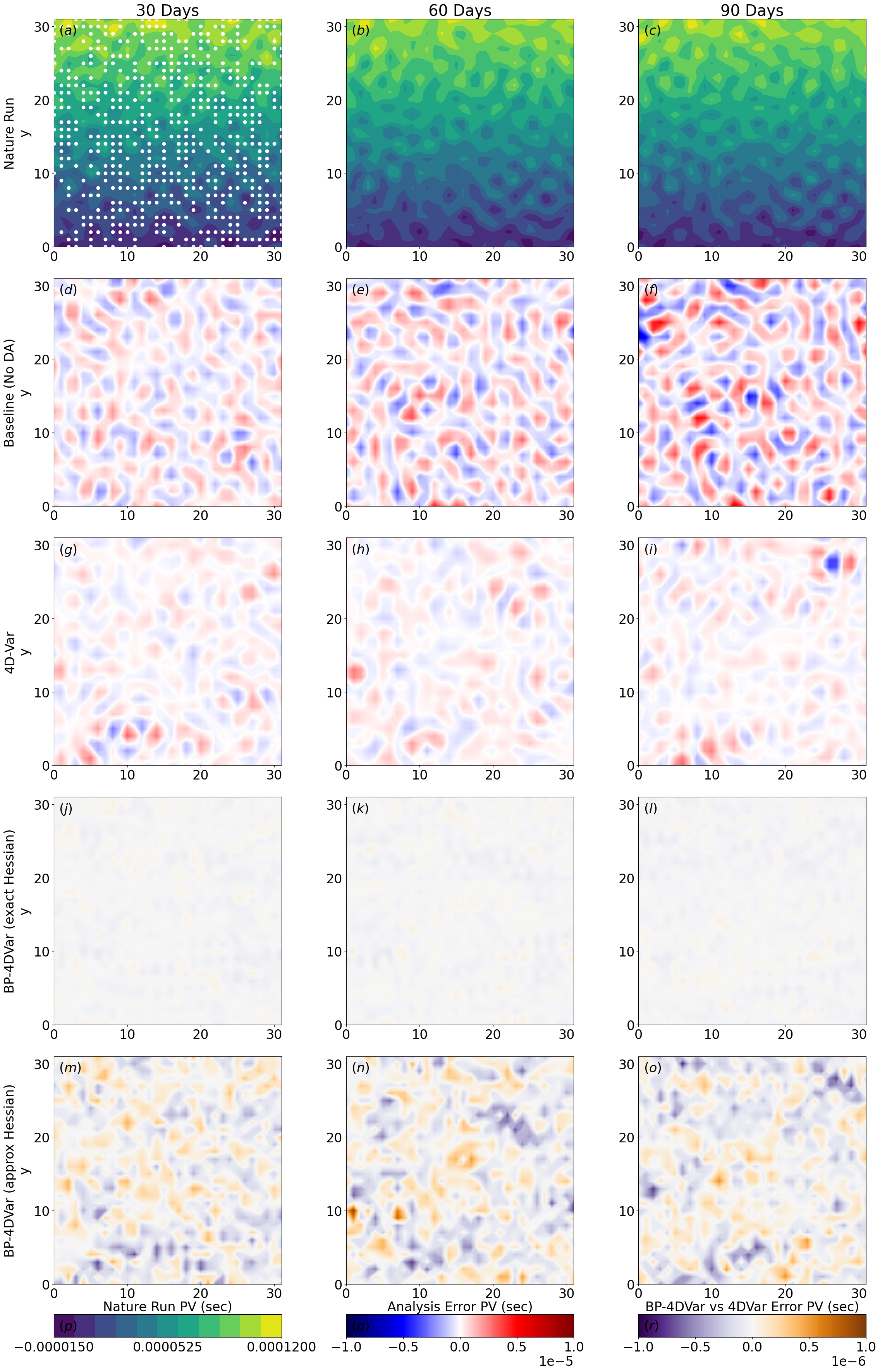}
\caption{PyQG experiment contour plots for the bottom layer of a 2048D system (2x32x32 in gridded space) at 3 times during the test period. The top row (a-c) shows the nature run and the observation locations in white, which are randomly sampled at half of the grid locations at every 3rd timestep. The next two rows show the error vs. the nature run for a baseline model run with no data assimilation (d-f) and for 4D-Var (g-i). Finally, the two variants of Backprop-4DVar---with the exact Hessian (j-l) and the approximate Hessian (m-o)---are shown in comparison to 4D-Var. Purple and orange represent lower and higher absolute error than 4D-Var respectively.}
\label{pyqg_jax_progression}
\end{center}
\end{figure}

In Figure \ref{pyqg_jax_progression} we show snapshots of the evolution of the lower layer of the 2048D (2x32x32) PyQG system at 30, 60, and 90 days (1,080 model time steps) into the test period. Using 4D-Var to serve as a reference for comparison, we examine the results using Backprop-4DVar with both an exact and approximate Hessian. When using the exact Hessian, state estimates calculated using Backprop-4DVar are nearly identical to the conventional incremental 4D-Var, as we would expect. Using the much faster Backprop-4DVar with an approximate Hessian, the solution differs, with Backprop-4DVar producing lower error in some regions and higher in others. 

\begin{figure}
\noindent\includegraphics[width=\textwidth]{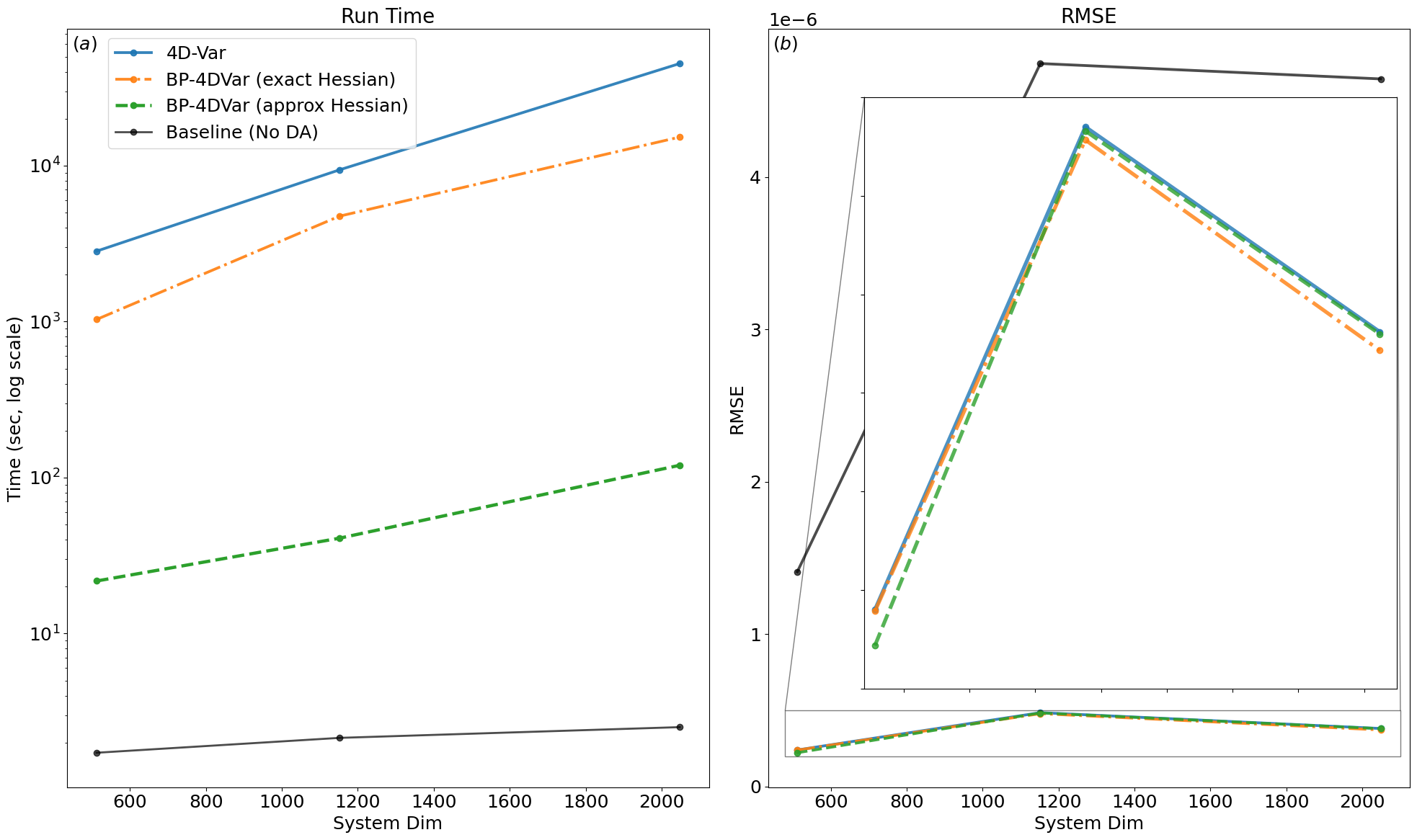}
\caption{(a) Run times (log scale) and (b) RMSE for the QG dynamics using the PyQG-JAX forecast model, for 4D-Var and Backprop-4DVar. An unconstrained free run without data assimilation is provided as a baseline for comparison. While all three DA methods show similar performance in terms of RMSE, Backprop-4DVar using the approximate Hessian (green) is an order of magnitude faster than the reference methods.}
\label{pyqg_da_results}
\end{figure}

Overall, the total RMSE over the full test period is very similar between the three DA methods. Figure \ref{pyqg_da_results} shows the run times and RMSE for 4D-Var (using 3 outer loops) and Backprop 4D-Var (using 3 iterations) with two different methods of computing the Hessian, while varying the system dimension. All of the DA methods produce state estimates with similar RMSE. Notably, however, the Backprop-4DVar with the approximate Hessian has much lower run times than the other methods, which is particularly apparent as the system size increases.

\begin{figure}
\begin{center}
\noindent\includegraphics[width=\textwidth,keepaspectratio]{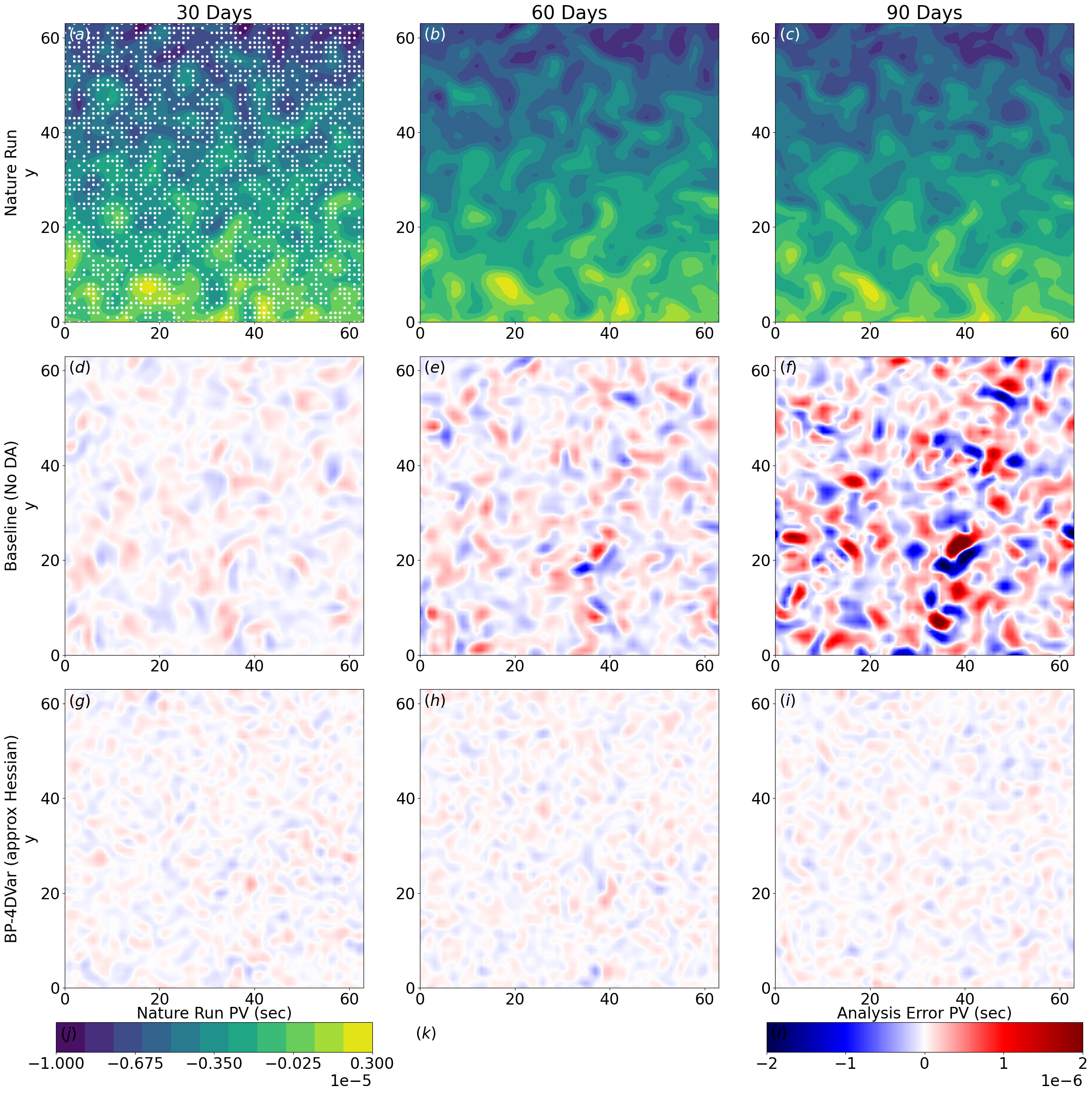}
\caption{Contour plots of the top layer of a 8192D PyQG system (2x64x64 in gridded space) for the (top row, a-c) nature run, (middle row, d-f) error for a baseline free run without DA, and (bottom row, g-i) error of the Backprop-4DVar with the approximate hessian. Observations are randomly sampled at half of all grid locations and taken every 3rd time step, and shown overlaid on the nature run in white.}
\label{pyqg_jax_64d}
\end{center}
\end{figure}

We conclude this section by applying Backprop-4DVar to a higher resolution 8192D (or 2x64x64) test case. Figure \ref{pyqg_jax_64d} shows the nature run integration, the location of observations used for the DA, the error of a free run initialized from a perturbed initial condition, and the error of the Backprop-4DVar with the approximate Hessian initiated from the same initial conditions. At this resolution, Backprop-4DVar has similar computational performance as shown in Figure \ref{pyqg_da_results}, while producing an RMSE of $3.43\times10^{-7}$ aggregated over the test set. For comparison, the RMSE for the 2048D system shown in Figure \ref{pyqg_da_results} is $3.80\times10^{-7}$.

\subsection{Quasi-geostrophic dynamics emulated with a Reservoir Computing Model}

\begin{figure}
\noindent\includegraphics[width=\textwidth]{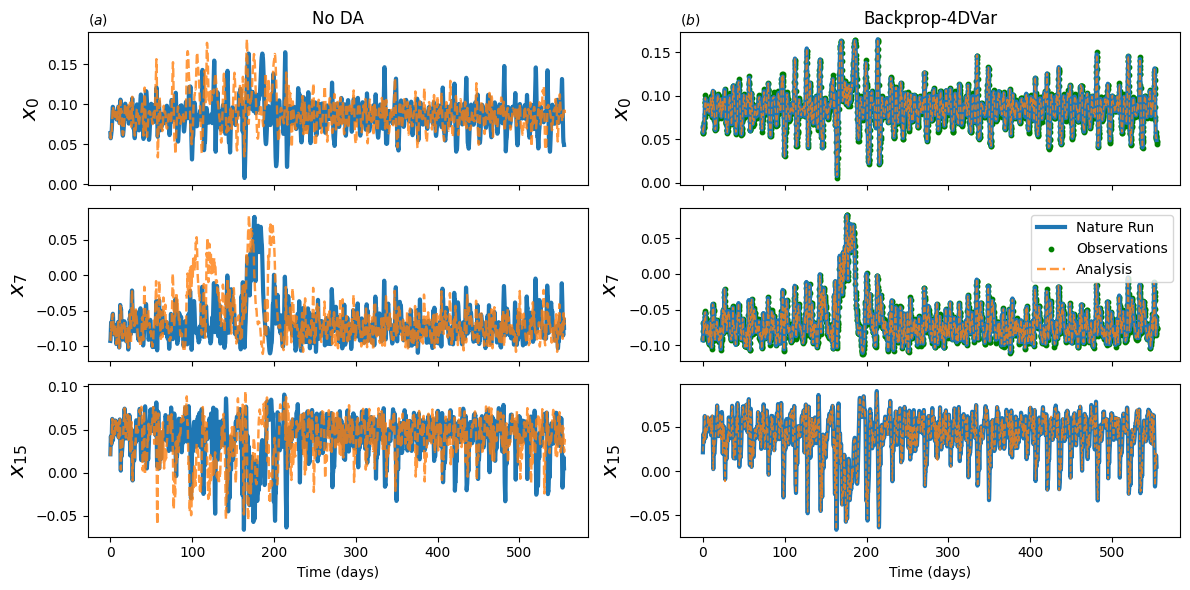}
\caption{The reconstructed trajectories of two observed ($x_0$ and $x_7$) and one unobserved ($x_{15}$) components of the 20D quasi-geostrophic dynamics constrained by Backprop-4DVar, using a surrogate reservoir computing forecast model, with 10 out of 20 variables observed. The standard deviation of the observation noise  is 10\% of the per-variable climatological standard deviation (calculated from the training period), and the forecast error covariance is a diagonal matrix with values equal to the per-variable climatological standard deviation in the reservoir space multiplied by 0.1. Here, 3 out of 20 variables are shown in panel (b) to illustrate the performance of the DA algorithm in comparison with the unconstrained model run shown in panel (a).}
\label{qgs_da_results}
\end{figure}

Finally, we apply the Backprop-4DVar in the case that a software-differentiable form of the system of equations is not available. In our case, we are using an external software package that would take considerable effort to rewrite in JAX. Instead, we use reservoir computing to produce a surrogate model that estimates the forward dynamics in Backprop-4DVar. We find that even when using this surrogate RC model in place of the `perfect' numerical model, Backprop-4DVar is still able to accurately reconstruct the trajectory of the QG baroclinic dynamics despite the noisy and sparse observations. The results of the DA experiment for 3 of the 20 model variables are shown in Figure \ref{qgs_da_results} in comparison with an unconstrained model run (without data assimilation). The RMSE for the 10,000 model time steps in the test period is 0.00529. 

\subsection{Learning Rate Tuning}

The performance of Backprop-4DVar with the approximate Hessian is sensitive to the learning rate ($\alpha$) and learning rate decay. If the learning rate is too low the DA cycle does not converge, eventually leading to filter divergence. If the learning rate is too high then the backpropagation runs the risk of producing exploding gradients. Our recommendation is to conduct a Bayesian optimization search for the learning rate and learning rate decay using the validation set (or a subset of it for computationally demanding experiments) using RayTune or a similar tool. The full results of the Bayesian optimization search for learning rate and learning rate decay are provided in the Supporting Information. In general, a learning rate of 0.5 and learning rate decay of 0.5 appear to perform decently well on most systems, but for the qgs experiments a much smaller learning rate performs better. For the qgs experiment, a learning rate of 0.019 and decay of 0.875 are found to perform best. For the Backprop-4DVar with the exact Hessian we use an $\alpha$ of 1 for all iterations.

\section{Conclusion}

We developed a new approach for performing 4D-variational (4D-Var) data assimilation (DA), called Backprop-4DVar, by implementing the Gauss-Newton method using a Hessian approximation along with machine learning tools that perform automatic differentiation and backpropogation. The Backprop-4DVar approach can be applied with any weather forecast model that supports automatic differentiation, whether it be a conventional numerical model (which we demonstrated with Lorenz 96 and PyQG dynamics) or a machine learning model (which we demonstrated with a reservoir computing surrogate model for quasi-geostrophic dynamics). In all of the Backprop-4DVar examples, the JAX Autograd autodifferentation capability was leveraged to calculate the necessary gradients, thus obviating the need to develop a tangent linear or adjoint model.

The runtime of Backprop-4DVar scaled nearly linearly with increasing system size, indicating its potential for use with much higher dimensional systems than tested here. The desired forecast skill and computational cost of the Backprop-4DVar approach are tunable via the learning rate and number of iterations (or `outer loops') per DA analysis cycle. We provided some general recommendations as starting points ($iterations \approx 3$, $\alpha \approx 0.5$, $\alpha_{decay} \approx 0.5$) and an approach for optimizing learning rate and learning rate decay using hyperparameter tuning. 

Backprop-4DVar ran substantially faster than conventional incremental 4D-Var. We implemented each in JAX using similar programming logic to create as fair a comparison as possible, but it should be noted that operational 4D-Var implementations (e.g. the implementation used by ECMWF) use a variety of optimizations to improve the computational performance, such as introducing new observations at each outer loop rather than waiting for the full analysis cycle to complete \cite{lean_continuous_2021}. Nevertheless, even with these developments the computational cost of 4D-Var is very heavy in operational forecast systems and remains a limitation \cite{bonavita_4d-var_2021}. Additionally, we note that all experiments were run on a CPU. JAX can utilize GPUs and TPUs for increased performance speed and the effect of this on Backprop-4DVar was not evaluated.

The learning rate and learning rate decay schedule are tuned using Bayesian optimization to automatically identify the optimal parameter values prior to commencing DA experiments. This step is recommended, since the direct gradient descent solver applied with Backprop-4DVar is sensitive to the learning rate: too large and it runs into the exploding gradients problem and RMSE increases rapidly, too small and it may not converge to reach a sufficiently accurate analysis. Future work will explore a range of nonlinear optimization strategies while expanding the scope to to global forecast models.

\section{Open Research}

Version 0.1.1-alpha of DataAssimBench used to perform all analyses is preserved at doi:10.5281/zenodo.13158498, available via the Apache-2.0 license and developed openly at https://github.com/StevePny/DataAssimBench. \cite{solvik_stevepnydataassimbench_2024}.

Version 0.1.1-alpha of DataAssimBench-Examples, containing the Jupyter Notebooks used to execute the analyses and create all figures, is preserved at \break doi:10.5281/zenodo.13158546, available via the Apache-2.0 license and developed openly at https://github.com/StevePny/DataAssimBench-Examples. \cite{solvik_stevepnydataassimbench-examples_2024}.


\acknowledgments
The authors acknowledge a gift from Google that supported this work. SGP proposed the research idea and methods. KS and SGP carried out the implementation of the methods. SH proposed the use of JAX automatic differentiation, improvements to the experiment design, and provided editorial guidance.


%
\bibliography{main}

\begin{thebibliography}{}

\bibitem [\protect \citeauthoryear {%
Abarbanel%
, Kostuk%
\BCBL {}\ \BBA {} Whartenby%
}{%
Abarbanel%
\ \protect \BOthers {.}}{%
{\protect \APACyear {2010}}%
}]{%
Abarbanel2010}
\APACinsertmetastar {%
Abarbanel2010}%
\begin{APACrefauthors}%
Abarbanel, H\BPBI D\BPBI I.%
, Kostuk, M.%
\BCBL {}\ \BBA {} Whartenby, W.%
\end{APACrefauthors}%
\unskip\
\newblock
\APACrefYearMonthDay{2010}{}{}.
\newblock
{\BBOQ}\APACrefatitle {Data assimilation with regularized nonlinear instabilities} {Data assimilation with regularized nonlinear instabilities}.{\BBCQ}
\newblock
\APACjournalVolNumPages{Quarterly Journal of the Royal Meteorological Society}{136}{648}{769-783}.
\newblock
\begin{APACrefURL} \url{https://rmets.onlinelibrary.wiley.com/doi/abs/10.1002/qj.600} \end{APACrefURL}
\newblock
\begin{APACrefDOI} \doi{https://doi.org/10.1002/qj.600} \end{APACrefDOI}
\PrintBackRefs{\CurrentBib}

\bibitem [\protect \citeauthoryear {%
Abernathay%
\ \BBA {} Ross%
}{%
Abernathay%
\ \BBA {} Ross%
}{%
{\protect \APACyear {2022}}%
}]{%
abernathay_pyqg_2022}
\APACinsertmetastar {%
abernathay_pyqg_2022}%
\begin{APACrefauthors}%
Abernathay, R.%
\BCBT {}\ \BBA {} Ross, A.%
\end{APACrefauthors}%
\unskip\
\newblock
\APACrefYearMonthDay{2022}{}{}.
\newblock
\APACrefbtitle {pyqg: python quasigeostrophic model.} {pyqg: python quasigeostrophic model.}
\newblock
\begin{APACrefURL} [{2024-04-12}]\url{http://github.com/pyqg/pyqg} \end{APACrefURL}
\PrintBackRefs{\CurrentBib}

\bibitem [\protect \citeauthoryear {%
Ansel%
\ \protect \BOthers {.}}{%
Ansel%
\ \protect \BOthers {.}}{%
{\protect \APACyear {2024}}%
}]{%
ansel_pytorch_2024}
\APACinsertmetastar {%
ansel_pytorch_2024}%
\begin{APACrefauthors}%
Ansel, J.%
, Yang, E.%
, He, H.%
, Gimelshein, N.%
, Jain, A.%
, Voznesensky, M.%
\BDBL {}Chintala, S.%
\end{APACrefauthors}%
\unskip\
\newblock
\APACrefYearMonthDay{2024}{{\APACmonth{04}}}{}.
\newblock
{\BBOQ}\APACrefatitle {{PyTorch} 2: {Faster} {Machine} {Learning} {Through} {Dynamic} {Python} {Bytecode} {Transformation} and {Graph} {Compilation}} {{PyTorch} 2: {Faster} {Machine} {Learning} {Through} {Dynamic} {Python} {Bytecode} {Transformation} and {Graph} {Compilation}}.{\BBCQ}
\newblock
\BIn{} \APACrefbtitle {Proceedings of the 29th {ACM} {International} {Conference} on {Architectural} {Support} for {Programming} {Languages} and {Operating} {Systems}, {Volume} 2} {Proceedings of the 29th {ACM} {International} {Conference} on {Architectural} {Support} for {Programming} {Languages} and {Operating} {Systems}, {Volume} 2}\ (\BVOL~2, \BPGS\ 929--947).
\newblock
\APACaddressPublisher{New York, NY, USA}{Association for Computing Machinery}.
\newblock
\begin{APACrefURL} [{2024-07-29}]\url{https://doi.org/10.1145/3620665.3640366} \end{APACrefURL}
\newblock
\begin{APACrefDOI} \doi{10.1145/3620665.3640366} \end{APACrefDOI}
\PrintBackRefs{\CurrentBib}

\bibitem [\protect \citeauthoryear {%
Bach%
\ \BBA {} Ghil%
}{%
Bach%
\ \BBA {} Ghil%
}{%
{\protect \APACyear {2023}}%
}]{%
bach2023}
\APACinsertmetastar {%
bach2023}%
\begin{APACrefauthors}%
Bach, E.%
\BCBT {}\ \BBA {} Ghil, M.%
\end{APACrefauthors}%
\unskip\
\newblock
\APACrefYearMonthDay{2023}{}{}.
\newblock
{\BBOQ}\APACrefatitle {A Multi-Model Ensemble Kalman Filter for Data Assimilation and Forecasting} {A multi-model ensemble kalman filter for data assimilation and forecasting}.{\BBCQ}
\newblock
\APACjournalVolNumPages{Journal of Advances in Modeling Earth Systems}{15}{1}{e2022MS003123}.
\newblock
\begin{APACrefURL} \url{https://agupubs.onlinelibrary.wiley.com/doi/abs/10.1029/2022MS003123} \end{APACrefURL}
\newblock
\APACrefnote{e2022MS003123 2022MS003123}
\newblock
\begin{APACrefDOI} \doi{https://doi.org/10.1029/2022MS003123} \end{APACrefDOI}
\PrintBackRefs{\CurrentBib}

\bibitem [\protect \citeauthoryear {%
Bergstra%
, Komer%
, Eliasmith%
, Yamins%
\BCBL {}\ \BBA {} Cox%
}{%
Bergstra%
\ \protect \BOthers {.}}{%
{\protect \APACyear {2015}}%
}]{%
Bergstra_2015}
\APACinsertmetastar {%
Bergstra_2015}%
\begin{APACrefauthors}%
Bergstra, J.%
, Komer, B.%
, Eliasmith, C.%
, Yamins, D.%
\BCBL {}\ \BBA {} Cox, D\BPBI D.%
\end{APACrefauthors}%
\unskip\
\newblock
\APACrefYearMonthDay{2015}{July}{}.
\newblock
{\BBOQ}\APACrefatitle {Hyperopt: a Python library for model selection and hyperparameter optimization} {Hyperopt: a python library for model selection and hyperparameter optimization}.{\BBCQ}
\newblock
\APACjournalVolNumPages{Computational Science \& Discovery}{8}{1}{014008}.
\newblock
\begin{APACrefURL} \url{https://dx.doi.org/10.1088/1749-4699/8/1/014008} \end{APACrefURL}
\newblock
\begin{APACrefDOI} \doi{10.1088/1749-4699/8/1/014008} \end{APACrefDOI}
\PrintBackRefs{\CurrentBib}

\bibitem [\protect \citeauthoryear {%
Bezanson%
, Edelman%
, Karpinski%
\BCBL {}\ \BBA {} Shah%
}{%
Bezanson%
\ \protect \BOthers {.}}{%
{\protect \APACyear {2017}}%
}]{%
Julia-2017}
\APACinsertmetastar {%
Julia-2017}%
\begin{APACrefauthors}%
Bezanson, J.%
, Edelman, A.%
, Karpinski, S.%
\BCBL {}\ \BBA {} Shah, V\BPBI B.%
\end{APACrefauthors}%
\unskip\
\newblock
\APACrefYearMonthDay{2017}{}{}.
\newblock
{\BBOQ}\APACrefatitle {Julia: A fresh approach to numerical computing} {Julia: A fresh approach to numerical computing}.{\BBCQ}
\newblock
\APACjournalVolNumPages{SIAM {R}eview}{59}{1}{65--98}.
\newblock
\begin{APACrefURL} \url{https://epubs.siam.org/doi/10.1137/141000671} \end{APACrefURL}
\newblock
\begin{APACrefDOI} \doi{10.1137/141000671} \end{APACrefDOI}
\PrintBackRefs{\CurrentBib}

\bibitem [\protect \citeauthoryear {%
Bi%
, Xie%
, Zhang%
\BCBL {}\ \BBA {} et al.%
}{%
Bi%
\ \protect \BOthers {.}}{%
{\protect \APACyear {2023}}%
}]{%
Bi2023}
\APACinsertmetastar {%
Bi2023}%
\begin{APACrefauthors}%
Bi, K.%
, Xie, L.%
, Zhang, H.%
\BCBL {}\ \BBA {} et al.%
\end{APACrefauthors}%
\unskip\
\newblock
\APACrefYearMonthDay{2023}{}{}.
\newblock
{\BBOQ}\APACrefatitle {Accurate medium-range global weather forecasting with 3D neural networks} {Accurate medium-range global weather forecasting with 3d neural networks}.{\BBCQ}
\newblock
\APACjournalVolNumPages{Nature}{619}{}{533–538}.
\PrintBackRefs{\CurrentBib}

\bibitem [\protect \citeauthoryear {%
Bonavita%
\ \BBA {} Lean%
}{%
Bonavita%
\ \BBA {} Lean%
}{%
{\protect \APACyear {2021}}%
}]{%
bonavita_4d-var_2021}
\APACinsertmetastar {%
bonavita_4d-var_2021}%
\begin{APACrefauthors}%
Bonavita, M.%
\BCBT {}\ \BBA {} Lean, P.%
\end{APACrefauthors}%
\unskip\
\newblock
\APACrefYearMonthDay{2021}{}{}.
\newblock
{\BBOQ}\APACrefatitle {{4D}-{Var} for numerical weather prediction} {{4D}-{Var} for numerical weather prediction}.{\BBCQ}
\newblock
\APACjournalVolNumPages{Weather}{76}{2}{65--66}.
\newblock
\begin{APACrefURL} [{2024-05-06}]\url{https://onlinelibrary.wiley.com/doi/abs/10.1002/wea.3862} \end{APACrefURL}
\newblock
\APACrefnote{\_eprint: https://onlinelibrary.wiley.com/doi/pdf/10.1002/wea.3862}
\newblock
\begin{APACrefDOI} \doi{10.1002/wea.3862} \end{APACrefDOI}
\PrintBackRefs{\CurrentBib}

\bibitem [\protect \citeauthoryear {%
Bonev%
\ \protect \BOthers {.}}{%
Bonev%
\ \protect \BOthers {.}}{%
{\protect \APACyear {2023}}%
}]{%
bonev_spherical_2023}
\APACinsertmetastar {%
bonev_spherical_2023}%
\begin{APACrefauthors}%
Bonev, B.%
, Kurth, T.%
, Hundt, C.%
, Pathak, J.%
, Baust, M.%
, Kashinath, K.%
\BCBL {}\ \BBA {} Anandkumar, A.%
\end{APACrefauthors}%
\unskip\
\newblock
\APACrefYearMonthDay{2023}{}{}.
\newblock
{\BBOQ}\APACrefatitle {Spherical Fourier neural operators: learning stable dynamics on the sphere} {Spherical fourier neural operators: learning stable dynamics on the sphere}.{\BBCQ}
\newblock
\BIn{} \APACrefbtitle {Proceedings of the 40th International Conference on Machine Learning.} {Proceedings of the 40th international conference on machine learning.}
\newblock
\APACaddressPublisher{}{JMLR.org}.
\PrintBackRefs{\CurrentBib}

\bibitem [\protect \citeauthoryear {%
Bradbury%
\ \protect \BOthers {.}}{%
Bradbury%
\ \protect \BOthers {.}}{%
{\protect \APACyear {2018}}%
}]{%
jax2018github}
\APACinsertmetastar {%
jax2018github}%
\begin{APACrefauthors}%
Bradbury, J.%
, Frostig, R.%
, Hawkins, P.%
, Johnson, M\BPBI J.%
, Leary, C.%
, Maclaurin, D.%
\BDBL {}Zhang, Q.%
\end{APACrefauthors}%
\unskip\
\newblock
\APACrefYearMonthDay{2018}{}{}.
\newblock
\APACrefbtitle {{JAX}: composable transformations of {P}ython+{N}um{P}y programs.} {{JAX}: composable transformations of {P}ython+{N}um{P}y programs.}
\newblock
\begin{APACrefURL} \url{http://github.com/google/jax} \end{APACrefURL}
\PrintBackRefs{\CurrentBib}

\bibitem [\protect \citeauthoryear {%
Carrassi%
, Ghil%
, Trevisan%
\BCBL {}\ \BBA {} Uboldi%
}{%
Carrassi%
\ \protect \BOthers {.}}{%
{\protect \APACyear {2008}}%
}]{%
carrassi2008}
\APACinsertmetastar {%
carrassi2008}%
\begin{APACrefauthors}%
Carrassi, A.%
, Ghil, M.%
, Trevisan, A.%
\BCBL {}\ \BBA {} Uboldi, F.%
\end{APACrefauthors}%
\unskip\
\newblock
\APACrefYearMonthDay{2008}{05}{}.
\newblock
{\BBOQ}\APACrefatitle {{Data assimilation as a nonlinear dynamical systems problem: Stability and convergence of the prediction-assimilation system}} {{Data assimilation as a nonlinear dynamical systems problem: Stability and convergence of the prediction-assimilation system}}.{\BBCQ}
\newblock
\APACjournalVolNumPages{Chaos: An Interdisciplinary Journal of Nonlinear Science}{18}{2}{023112}.
\newblock
\begin{APACrefURL} \url{https://doi.org/10.1063/1.2909862} \end{APACrefURL}
\newblock
\begin{APACrefDOI} \doi{10.1063/1.2909862} \end{APACrefDOI}
\PrintBackRefs{\CurrentBib}

\bibitem [\protect \citeauthoryear {%
Cartis%
, Kaouri%
, Lawless%
\BCBL {}\ \BBA {} Nichols%
}{%
Cartis%
\ \protect \BOthers {.}}{%
{\protect \APACyear {2021}}%
}]{%
cartis_convergent_2021}
\APACinsertmetastar {%
cartis_convergent_2021}%
\begin{APACrefauthors}%
Cartis, C.%
, Kaouri, M\BPBI H.%
, Lawless, A\BPBI S.%
\BCBL {}\ \BBA {} Nichols, N\BPBI K.%
\end{APACrefauthors}%
\unskip\
\newblock
\APACrefYearMonthDay{2021}{{\APACmonth{07}}}{}.
\newblock
\APACrefbtitle {Convergent least-squares optimisation methods for variational data assimilation.} {Convergent least-squares optimisation methods for variational data assimilation.}
\newblock
\APACaddressPublisher{}{arXiv}.
\newblock
\begin{APACrefURL} [{2024-01-11}]\url{http://arxiv.org/abs/2107.12361} \end{APACrefURL}
\newblock
\APACrefnote{arXiv:2107.12361 [math]}
\PrintBackRefs{\CurrentBib}

\bibitem [\protect \citeauthoryear {%
Demaeyer%
, Cruz%
\BCBL {}\ \BBA {} Vannitsem%
}{%
Demaeyer%
\ \protect \BOthers {.}}{%
{\protect \APACyear {2020}}%
}]{%
demaeyer_qgs_2020}
\APACinsertmetastar {%
demaeyer_qgs_2020}%
\begin{APACrefauthors}%
Demaeyer, J.%
, Cruz, L\BPBI D.%
\BCBL {}\ \BBA {} Vannitsem, S.%
\end{APACrefauthors}%
\unskip\
\newblock
\APACrefYearMonthDay{2020}{{\APACmonth{12}}}{}.
\newblock
{\BBOQ}\APACrefatitle {qgs: {A} flexible {Python} framework of reduced-order multiscale climate models} {qgs: {A} flexible {Python} framework of reduced-order multiscale climate models}.{\BBCQ}
\newblock
\APACjournalVolNumPages{Journal of Open Source Software}{5}{56}{2597}.
\newblock
\begin{APACrefURL} [{2023-10-11}]\url{https://joss.theoj.org/papers/10.21105/joss.02597} \end{APACrefURL}
\newblock
\begin{APACrefDOI} \doi{10.21105/joss.02597} \end{APACrefDOI}
\PrintBackRefs{\CurrentBib}

\bibitem [\protect \citeauthoryear {%
Dembo%
, Eisenstat%
\BCBL {}\ \BBA {} Steihaug%
}{%
Dembo%
\ \protect \BOthers {.}}{%
{\protect \APACyear {1982}}%
}]{%
dembo_inexact_1982}
\APACinsertmetastar {%
dembo_inexact_1982}%
\begin{APACrefauthors}%
Dembo, R\BPBI S.%
, Eisenstat, S\BPBI C.%
\BCBL {}\ \BBA {} Steihaug, T.%
\end{APACrefauthors}%
\unskip\
\newblock
\APACrefYearMonthDay{1982}{}{}.
\newblock
{\BBOQ}\APACrefatitle {Inexact {Newton} {Methods}} {Inexact {Newton} {Methods}}.{\BBCQ}
\newblock
\APACjournalVolNumPages{SIAM Journal on Numerical Analysis}{19}{2}{400--408}.
\newblock
\begin{APACrefURL} [{2024-01-11}]\url{https://epubs.siam.org/doi/10.1137/0719025} \end{APACrefURL}
\newblock
\APACrefnote{Publisher: Society for Industrial and Applied Mathematics}
\newblock
\begin{APACrefDOI} \doi{10.1137/0719025} \end{APACrefDOI}
\PrintBackRefs{\CurrentBib}

\bibitem [\protect \citeauthoryear {%
Dimet%
, Navon%
\BCBL {}\ \BBA {} Daescu%
}{%
Dimet%
\ \protect \BOthers {.}}{%
{\protect \APACyear {2002}}%
}]{%
dimet_second-order_2002}
\APACinsertmetastar {%
dimet_second-order_2002}%
\begin{APACrefauthors}%
Dimet, F\BHBI X\BPBI L.%
, Navon, I\BPBI M.%
\BCBL {}\ \BBA {} Daescu, D\BPBI N.%
\end{APACrefauthors}%
\unskip\
\newblock
\APACrefYearMonthDay{2002}{{\APACmonth{03}}}{}.
\newblock
{\BBOQ}\APACrefatitle {Second-{Order} {Information} in {Data} {Assimilation}} {Second-{Order} {Information} in {Data} {Assimilation}}.{\BBCQ}
\newblock
\APACjournalVolNumPages{Monthly Weather Review}{130}{3}{629--648}.
\newblock
\begin{APACrefURL} [{2024-01-11}]\url{https://journals.ametsoc.org/view/journals/mwre/130/3/1520-0493_2002_130_0629_soiida_2.0.co_2.xml} \end{APACrefURL}
\newblock
\APACrefnote{Publisher: American Meteorological Society Section: Monthly Weather Review}
\newblock
\begin{APACrefDOI} \doi{10.1175/1520-0493(2002)130<0629:SOIIDA>2.0.CO;2} \end{APACrefDOI}
\PrintBackRefs{\CurrentBib}

\bibitem [\protect \citeauthoryear {%
Dormand%
\ \BBA {} Prince%
}{%
Dormand%
\ \BBA {} Prince%
}{%
{\protect \APACyear {1980}}%
}]{%
dormand1980family}
\APACinsertmetastar {%
dormand1980family}%
\begin{APACrefauthors}%
Dormand, J\BPBI R.%
\BCBT {}\ \BBA {} Prince, P\BPBI J.%
\end{APACrefauthors}%
\unskip\
\newblock
\APACrefYearMonthDay{1980}{}{}.
\newblock
{\BBOQ}\APACrefatitle {A family of embedded Runge-Kutta formulae} {A family of embedded runge-kutta formulae}.{\BBCQ}
\newblock
\APACjournalVolNumPages{Journal of computational and applied mathematics}{6}{1}{19--26}.
\PrintBackRefs{\CurrentBib}

\bibitem [\protect \citeauthoryear {%
Fertig%
, Harlim%
\BCBL {}\ \BBA {} Hunt%
}{%
Fertig%
\ \protect \BOthers {.}}{%
{\protect \APACyear {2007}}%
}]{%
fertig2007}
\APACinsertmetastar {%
fertig2007}%
\begin{APACrefauthors}%
Fertig, E\BPBI J.%
, Harlim, J.%
\BCBL {}\ \BBA {} Hunt, B\BPBI R.%
\end{APACrefauthors}%
\unskip\
\newblock
\APACrefYearMonthDay{2007}{}{}.
\newblock
{\BBOQ}\APACrefatitle {A comparative study of 4D-VAR and a 4D Ensemble Kalman Filter: perfect model simulations with Lorenz-96} {A comparative study of 4d-var and a 4d ensemble kalman filter: perfect model simulations with lorenz-96}.{\BBCQ}
\newblock
\APACjournalVolNumPages{Tellus A: Dynamic Meteorology and Oceanography}{59}{1}{96-100}.
\newblock
\begin{APACrefURL} \url{https://doi.org/10.1111/j.1600-0870.2006.00205.x} \end{APACrefURL}
\newblock
\begin{APACrefDOI} \doi{10.1111/j.1600-0870.2006.00205.x} \end{APACrefDOI}
\PrintBackRefs{\CurrentBib}

\bibitem [\protect \citeauthoryear {%
Gratton%
, Lawless%
\BCBL {}\ \BBA {} Nichols%
}{%
Gratton%
\ \protect \BOthers {.}}{%
{\protect \APACyear {2007}}%
}]{%
gratton_approximate_2007}
\APACinsertmetastar {%
gratton_approximate_2007}%
\begin{APACrefauthors}%
Gratton, S.%
, Lawless, A\BPBI S.%
\BCBL {}\ \BBA {} Nichols, N\BPBI K.%
\end{APACrefauthors}%
\unskip\
\newblock
\APACrefYearMonthDay{2007}{{\APACmonth{01}}}{}.
\newblock
{\BBOQ}\APACrefatitle {Approximate {Gauss}–{Newton} {Methods} for {Nonlinear} {Least} {Squares} {Problems}} {Approximate {Gauss}–{Newton} {Methods} for {Nonlinear} {Least} {Squares} {Problems}}.{\BBCQ}
\newblock
\APACjournalVolNumPages{SIAM Journal on Optimization}{18}{1}{106--132}.
\newblock
\begin{APACrefURL} [{2024-01-11}]\url{https://epubs.siam.org/doi/10.1137/050624935} \end{APACrefURL}
\newblock
\APACrefnote{Publisher: Society for Industrial and Applied Mathematics}
\newblock
\begin{APACrefDOI} \doi{10.1137/050624935} \end{APACrefDOI}
\PrintBackRefs{\CurrentBib}

\bibitem [\protect \citeauthoryear {%
Hansen%
}{%
Hansen%
}{%
{\protect \APACyear {2023}}%
}]{%
hansen23}
\APACinsertmetastar {%
hansen23}%
\begin{APACrefauthors}%
Hansen, N.%
\end{APACrefauthors}%
\unskip\
\newblock
\APACrefYearMonthDay{2023}{}{}.
\newblock
\APACrefbtitle {The CMA Evolution Strategy: A Tutorial.} {The cma evolution strategy: A tutorial.}
\PrintBackRefs{\CurrentBib}

\bibitem [\protect \citeauthoryear {%
Hansen%
, Müller%
\BCBL {}\ \BBA {} Koumoutsakos%
}{%
Hansen%
\ \protect \BOthers {.}}{%
{\protect \APACyear {2003}}%
}]{%
hansen2003}
\APACinsertmetastar {%
hansen2003}%
\begin{APACrefauthors}%
Hansen, N.%
, Müller, S\BPBI D.%
\BCBL {}\ \BBA {} Koumoutsakos, P.%
\end{APACrefauthors}%
\unskip\
\newblock
\APACrefYearMonthDay{2003}{03}{}.
\newblock
{\BBOQ}\APACrefatitle {{Reducing the Time Complexity of the Derandomized Evolution Strategy with Covariance Matrix Adaptation (CMA-ES)}} {{Reducing the Time Complexity of the Derandomized Evolution Strategy with Covariance Matrix Adaptation (CMA-ES)}}.{\BBCQ}
\newblock
\APACjournalVolNumPages{Evolutionary Computation}{11}{1}{1-18}.
\newblock
\begin{APACrefURL} \url{https://doi.org/10.1162/106365603321828970} \end{APACrefURL}
\newblock
\begin{APACrefDOI} \doi{10.1162/106365603321828970} \end{APACrefDOI}
\PrintBackRefs{\CurrentBib}

\bibitem [\protect \citeauthoryear {%
Jaeger%
}{%
Jaeger%
}{%
{\protect \APACyear {2002}}%
}]{%
Jaeger02}
\APACinsertmetastar {%
Jaeger02}%
\begin{APACrefauthors}%
Jaeger, H.%
\end{APACrefauthors}%
\unskip\
\newblock
\APACrefYearMonthDay{2002}{}{}.
\newblock
{\BBOQ}\APACrefatitle {Short term memory in echo state networks. GMD-Report 152} {Short term memory in echo state networks. gmd-report 152}.{\BBCQ}
\newblock
\BIn{} \APACrefbtitle {GMD - German National Research Institute for Computer Science.} {Gmd - german national research institute for computer science.}
\PrintBackRefs{\CurrentBib}

\bibitem [\protect \citeauthoryear {%
Jaeger%
}{%
Jaeger%
}{%
{\protect \APACyear {2010}}%
}]{%
Jaeger01}
\APACinsertmetastar {%
Jaeger01}%
\begin{APACrefauthors}%
Jaeger, H.%
\end{APACrefauthors}%
\unskip\
\newblock
\APACrefYearMonthDay{2010}{}{}.
\newblock
{\BBOQ}\APACrefatitle {The "echo state" approach to analysing and training recurrent neural networks-with an erratum note} {The "echo state" approach to analysing and training recurrent neural networks-with an erratum note}.{\BBCQ}
\newblock
\APACjournalVolNumPages{Bonn, Germany: German National Research Center for Information Technology GMD Technical Report}{148}{}{1-47}.
\PrintBackRefs{\CurrentBib}

\bibitem [\protect \citeauthoryear {%
Jaeger%
}{%
Jaeger%
}{%
{\protect \APACyear {2012}}%
}]{%
Jaeger12}
\APACinsertmetastar {%
Jaeger12}%
\begin{APACrefauthors}%
Jaeger, H.%
\end{APACrefauthors}%
\unskip\
\newblock
\APACrefYearMonthDay{2012}{}{}.
\newblock
\APACrefbtitle {Long Short-Term Memory in Echo State Networks: Details of a Simulation Study} {Long short-term memory in echo state networks: Details of a simulation study}\ (\BNUM~27).
\PrintBackRefs{\CurrentBib}

\bibitem [\protect \citeauthoryear {%
Jaeger%
\ \BBA {} Haas%
}{%
Jaeger%
\ \BBA {} Haas%
}{%
{\protect \APACyear {2004}}%
}]{%
Jaeger04}
\APACinsertmetastar {%
Jaeger04}%
\begin{APACrefauthors}%
Jaeger, H.%
\BCBT {}\ \BBA {} Haas, H.%
\end{APACrefauthors}%
\unskip\
\newblock
\APACrefYearMonthDay{2004}{}{}.
\newblock
{\BBOQ}\APACrefatitle {Harnessing Nonlinearity: Predicting Chaotic Systems and Saving Energy in Wireless Communication} {Harnessing nonlinearity: Predicting chaotic systems and saving energy in wireless communication}.{\BBCQ}
\newblock
\APACjournalVolNumPages{Science}{304}{}{78--80}.
\newblock
\begin{APACrefDOI} \doi{10.1126/science.1091277} \end{APACrefDOI}
\PrintBackRefs{\CurrentBib}

\bibitem [\protect \citeauthoryear {%
Kochkov%
\ \protect \BOthers {.}}{%
Kochkov%
\ \protect \BOthers {.}}{%
{\protect \APACyear {2023}}%
}]{%
kochkov2023neural}
\APACinsertmetastar {%
kochkov2023neural}%
\begin{APACrefauthors}%
Kochkov, D.%
, Yuval, J.%
, Langmore, I.%
, Norgaard, P.%
, Smith, J.%
, Mooers, G.%
\BDBL {}Hoyer, S.%
\end{APACrefauthors}%
\unskip\
\newblock
\APACrefYearMonthDay{2023}{}{}.
\newblock
{\BBOQ}\APACrefatitle {Neural General Circulation Models} {Neural general circulation models}.{\BBCQ}
\newblock

\PrintBackRefs{\CurrentBib}

\bibitem [\protect \citeauthoryear {%
Kurth%
\ \protect \BOthers {.}}{%
Kurth%
\ \protect \BOthers {.}}{%
{\protect \APACyear {2023}}%
}]{%
kurth2023}
\APACinsertmetastar {%
kurth2023}%
\begin{APACrefauthors}%
Kurth, T.%
, Subramanian, S.%
, Harrington, P.%
, Pathak, J.%
, Mardani, M.%
, Hall, D.%
\BDBL {}Anandkumar, A.%
\end{APACrefauthors}%
\unskip\
\newblock
\APACrefYearMonthDay{2023}{}{}.
\newblock
{\BBOQ}\APACrefatitle {FourCastNet: Accelerating Global High-Resolution Weather Forecasting Using Adaptive Fourier Neural Operators} {Fourcastnet: Accelerating global high-resolution weather forecasting using adaptive fourier neural operators}.{\BBCQ}
\newblock
\BIn{} \APACrefbtitle {Proceedings of the Platform for Advanced Scientific Computing Conference.} {Proceedings of the platform for advanced scientific computing conference.}
\newblock
\APACaddressPublisher{New York, NY, USA}{Association for Computing Machinery}.
\newblock
\begin{APACrefURL} \url{https://doi.org/10.1145/3592979.3593412} \end{APACrefURL}
\newblock
\begin{APACrefDOI} \doi{10.1145/3592979.3593412} \end{APACrefDOI}
\PrintBackRefs{\CurrentBib}

\bibitem [\protect \citeauthoryear {%
Lam%
\ \protect \BOthers {.}}{%
Lam%
\ \protect \BOthers {.}}{%
{\protect \APACyear {2023}}%
}]{%
lam2023}
\APACinsertmetastar {%
lam2023}%
\begin{APACrefauthors}%
Lam, R.%
, Sanchez-Gonzalez, A.%
, Willson, M.%
, Wirnsberger, P.%
, Fortunato, M.%
, Alet, F.%
\BDBL {}Battaglia, P.%
\end{APACrefauthors}%
\unskip\
\newblock
\APACrefYearMonthDay{2023}{}{}.
\newblock
{\BBOQ}\APACrefatitle {Learning skillful medium-range global weather forecasting} {Learning skillful medium-range global weather forecasting}.{\BBCQ}
\newblock
\APACjournalVolNumPages{Science}{0}{0}{eadi2336}.
\newblock
\begin{APACrefURL} \url{https://www.science.org/doi/abs/10.1126/science.adi2336} \end{APACrefURL}
\newblock
\begin{APACrefDOI} \doi{10.1126/science.adi2336} \end{APACrefDOI}
\PrintBackRefs{\CurrentBib}

\bibitem [\protect \citeauthoryear {%
Lawless%
, Gratton%
\BCBL {}\ \BBA {} Nichols%
}{%
Lawless%
\ \protect \BOthers {.}}{%
{\protect \APACyear {2005}}%
}]{%
lawless2005}
\APACinsertmetastar {%
lawless2005}%
\begin{APACrefauthors}%
Lawless, A\BPBI S.%
, Gratton, S.%
\BCBL {}\ \BBA {} Nichols, N\BPBI K.%
\end{APACrefauthors}%
\unskip\
\newblock
\APACrefYearMonthDay{2005}{}{}.
\newblock
{\BBOQ}\APACrefatitle {An investigation of incremental 4D-Var using non-tangent linear models} {An investigation of incremental 4d-var using non-tangent linear models}.{\BBCQ}
\newblock
\APACjournalVolNumPages{Quarterly Journal of the Royal Meteorological Society}{131}{606}{459-476}.
\newblock
\begin{APACrefURL} \url{https://rmets.onlinelibrary.wiley.com/doi/abs/10.1256/qj.04.20} \end{APACrefURL}
\newblock
\begin{APACrefDOI} \doi{https://doi.org/10.1256/qj.04.20} \end{APACrefDOI}
\PrintBackRefs{\CurrentBib}

\bibitem [\protect \citeauthoryear {%
Lawless%
\ \BBA {} Nichols%
}{%
Lawless%
\ \BBA {} Nichols%
}{%
{\protect \APACyear {2006}}%
}]{%
lawless_inner-loop_2006}
\APACinsertmetastar {%
lawless_inner-loop_2006}%
\begin{APACrefauthors}%
Lawless, A\BPBI S.%
\BCBT {}\ \BBA {} Nichols, N\BPBI K.%
\end{APACrefauthors}%
\unskip\
\newblock
\APACrefYearMonthDay{2006}{}{}.
\newblock
{\BBOQ}\APACrefatitle {Inner-{Loop} {Stopping} {Criteria} for {Incremental} {Four}-{Dimensional} {Variational} {Data} {Assimilation}} {Inner-{Loop} {Stopping} {Criteria} for {Incremental} {Four}-{Dimensional} {Variational} {Data} {Assimilation}}.{\BBCQ}
\newblock
\APACjournalVolNumPages{Monthly Weather Review}{134}{11}{3425--3435}.
\newblock
\begin{APACrefURL} [{2024-01-11}]\url{https://journals.ametsoc.org/view/journals/mwre/134/11/mwr3242.1.xml} \end{APACrefURL}
\newblock
\APACrefnote{Publisher: American Meteorological Society Section: Monthly Weather Review}
\newblock
\begin{APACrefDOI} \doi{10.1175/MWR3242.1} \end{APACrefDOI}
\PrintBackRefs{\CurrentBib}

\bibitem [\protect \citeauthoryear {%
Lean%
\ \protect \BOthers {.}}{%
Lean%
\ \protect \BOthers {.}}{%
{\protect \APACyear {2021}}%
}]{%
lean_continuous_2021}
\APACinsertmetastar {%
lean_continuous_2021}%
\begin{APACrefauthors}%
Lean, P.%
, Hólm, E\BPBI V.%
, Bonavita, M.%
, Bormann, N.%
, McNally, A\BPBI P.%
\BCBL {}\ \BBA {} Järvinen, H.%
\end{APACrefauthors}%
\unskip\
\newblock
\APACrefYearMonthDay{2021}{}{}.
\newblock
{\BBOQ}\APACrefatitle {Continuous data assimilation for global numerical weather prediction} {Continuous data assimilation for global numerical weather prediction}.{\BBCQ}
\newblock
\APACjournalVolNumPages{Quarterly Journal of the Royal Meteorological Society}{147}{734}{273--288}.
\newblock
\begin{APACrefURL} [{2024-05-06}]\url{https://onlinelibrary.wiley.com/doi/abs/10.1002/qj.3917} \end{APACrefURL}
\newblock
\APACrefnote{\_eprint: https://onlinelibrary.wiley.com/doi/pdf/10.1002/qj.3917}
\newblock
\begin{APACrefDOI} \doi{10.1002/qj.3917} \end{APACrefDOI}
\PrintBackRefs{\CurrentBib}

\bibitem [\protect \citeauthoryear {%
Liaw%
\ \protect \BOthers {.}}{%
Liaw%
\ \protect \BOthers {.}}{%
{\protect \APACyear {2018}}%
}]{%
liaw2018tune}
\APACinsertmetastar {%
liaw2018tune}%
\begin{APACrefauthors}%
Liaw, R.%
, Liang, E.%
, Nishihara, R.%
, Moritz, P.%
, Gonzalez, J\BPBI E.%
\BCBL {}\ \BBA {} Stoica, I.%
\end{APACrefauthors}%
\unskip\
\newblock
\APACrefYearMonthDay{2018}{}{}.
\newblock
\APACrefbtitle {Tune: A Research Platform for Distributed Model Selection and Training.} {Tune: A research platform for distributed model selection and training.}
\PrintBackRefs{\CurrentBib}

\bibitem [\protect \citeauthoryear {%
Lorenc%
}{%
Lorenc%
}{%
{\protect \APACyear {2000}}%
}]{%
Lorenc_2000}
\APACinsertmetastar {%
Lorenc_2000}%
\begin{APACrefauthors}%
Lorenc, A\BPBI C.%
\end{APACrefauthors}%
\unskip\
\newblock
\APACrefYearMonthDay{2000}{}{}.
\newblock
\APACrefbtitle {The Met. Office global 3-dimensional variational data assimilation scheme.} {The met. office global 3-dimensional variational data assimilation scheme.}\ (\BVOL~126).
\PrintBackRefs{\CurrentBib}

\bibitem [\protect \citeauthoryear {%
Lorenz%
}{%
Lorenz%
}{%
{\protect \APACyear {1996}}%
}]{%
lorenz1996predictability}
\APACinsertmetastar {%
lorenz1996predictability}%
\begin{APACrefauthors}%
Lorenz, E\BPBI N.%
\end{APACrefauthors}%
\unskip\
\newblock
\APACrefYearMonthDay{1996}{}{}.
\newblock
{\BBOQ}\APACrefatitle {Predictability: A problem partly solved} {Predictability: A problem partly solved}.{\BBCQ}
\newblock
\BIn{} \APACrefbtitle {Proc. Seminar on predictability} {Proc. seminar on predictability}\ (\BVOL~1).
\PrintBackRefs{\CurrentBib}

\bibitem [\protect \citeauthoryear {%
Lukosevicius%
}{%
Lukosevicius%
}{%
{\protect \APACyear {2012}}%
}]{%
Luko12}
\APACinsertmetastar {%
Luko12}%
\begin{APACrefauthors}%
Lukosevicius, M.%
\end{APACrefauthors}%
\unskip\
\newblock
\APACrefYearMonthDay{2012}{}{}.
\newblock
{\BBOQ}\APACrefatitle {A Practical Guide to Applying Echo State Networks} {A practical guide to applying echo state networks}.{\BBCQ}
\newblock
\APACjournalVolNumPages{}{}{}{659-686}.
\newblock
\begin{APACrefDOI} \doi{10.1007/978-3-642-35289-8_36} \end{APACrefDOI}
\PrintBackRefs{\CurrentBib}

\bibitem [\protect \citeauthoryear {%
Maass%
, Natschl{\"a}ger%
\BCBL {}\ \BBA {} Markram%
}{%
Maass%
\ \protect \BOthers {.}}{%
{\protect \APACyear {2002}}%
}]{%
Maass02}
\APACinsertmetastar {%
Maass02}%
\begin{APACrefauthors}%
Maass, W.%
, Natschl{\"a}ger, T.%
\BCBL {}\ \BBA {} Markram, H.%
\end{APACrefauthors}%
\unskip\
\newblock
\APACrefYearMonthDay{2002}{}{}.
\newblock
{\BBOQ}\APACrefatitle {Real-Time Computing Without Stable States: A New Framework for Neural Computation Based on Perturbations} {Real-time computing without stable states: A new framework for neural computation based on perturbations}.{\BBCQ}
\newblock
\APACjournalVolNumPages{Neural Computation}{14}{}{2531-2560}.
\PrintBackRefs{\CurrentBib}

\bibitem [\protect \citeauthoryear {%
Mahfouf%
\ \BBA {} Rabier%
}{%
Mahfouf%
\ \BBA {} Rabier%
}{%
{\protect \APACyear {2000}}%
}]{%
Mahfouf_2000}
\APACinsertmetastar {%
Mahfouf_2000}%
\begin{APACrefauthors}%
Mahfouf, J\BHBI F.%
\BCBT {}\ \BBA {} Rabier, F.%
\end{APACrefauthors}%
\unskip\
\newblock
\APACrefYearMonthDay{2000}{}{}.
\newblock
\APACrefbtitle {The ECMWF operational implementation of four dimensional variational assimilation. II: Experimental results with improved physics.} {The ecmwf operational implementation of four dimensional variational assimilation. ii: Experimental results with improved physics.}\ (\BVOL~126).
\PrintBackRefs{\CurrentBib}

\bibitem [\protect \citeauthoryear {%
McWilliams%
}{%
McWilliams%
}{%
{\protect \APACyear {1984}}%
}]{%
mcwilliams_emergence_1984}
\APACinsertmetastar {%
mcwilliams_emergence_1984}%
\begin{APACrefauthors}%
McWilliams, J\BPBI C.%
\end{APACrefauthors}%
\unskip\
\newblock
\APACrefYearMonthDay{1984}{{\APACmonth{09}}}{}.
\newblock
{\BBOQ}\APACrefatitle {The emergence of isolated coherent vortices in turbulent flow} {The emergence of isolated coherent vortices in turbulent flow}.{\BBCQ}
\newblock
\APACjournalVolNumPages{Journal of Fluid Mechanics}{146}{}{21--43}.
\newblock
\begin{APACrefURL} [{2024-04-12}]\url{https://www.cambridge.org/core/journals/journal-of-fluid-mechanics/article/abs/emergence-of-isolated-coherent-vortices-in-turbulent-flow/3EB789299B1A730265A5EC522E35B630} \end{APACrefURL}
\newblock
\begin{APACrefDOI} \doi{10.1017/S0022112084001750} \end{APACrefDOI}
\PrintBackRefs{\CurrentBib}

\bibitem [\protect \citeauthoryear {%
Nerger%
}{%
Nerger%
}{%
{\protect \APACyear {2022}}%
}]{%
nerger2022}
\APACinsertmetastar {%
nerger2022}%
\begin{APACrefauthors}%
Nerger, L.%
\end{APACrefauthors}%
\unskip\
\newblock
\APACrefYearMonthDay{2022}{}{}.
\newblock
{\BBOQ}\APACrefatitle {Data assimilation for nonlinear systems with a hybrid nonlinear Kalman ensemble transform filter} {Data assimilation for nonlinear systems with a hybrid nonlinear kalman ensemble transform filter}.{\BBCQ}
\newblock
\APACjournalVolNumPages{Quarterly Journal of the Royal Meteorological Society}{148}{743}{620-640}.
\newblock
\begin{APACrefURL} \url{https://rmets.onlinelibrary.wiley.com/doi/abs/10.1002/qj.4221} \end{APACrefURL}
\newblock
\begin{APACrefDOI} \doi{https://doi.org/10.1002/qj.4221} \end{APACrefDOI}
\PrintBackRefs{\CurrentBib}

\bibitem [\protect \citeauthoryear {%
Otness%
}{%
Otness%
}{%
{\protect \APACyear {2024}}%
}]{%
otness_pyqg-jax_2024}
\APACinsertmetastar {%
otness_pyqg-jax_2024}%
\begin{APACrefauthors}%
Otness, K.%
\end{APACrefauthors}%
\unskip\
\newblock
\APACrefYearMonthDay{2024}{}{}.
\newblock
\APACrefbtitle {pyqg-jax: {Quasigeostrophic} model in {JAX} (port of {PyQG}).} {pyqg-jax: {Quasigeostrophic} model in {JAX} (port of {PyQG}).}
\newblock
\begin{APACrefURL} [{2024-04-12}]\url{https://github.com/karlotness/pyqg-jax} \end{APACrefURL}
\PrintBackRefs{\CurrentBib}

\bibitem [\protect \citeauthoryear {%
Penny%
}{%
Penny%
}{%
{\protect \APACyear {2014}}%
}]{%
Penny2014}
\APACinsertmetastar {%
Penny2014}%
\begin{APACrefauthors}%
Penny, S\BPBI G.%
\end{APACrefauthors}%
\unskip\
\newblock
\APACrefYearMonthDay{2014}{}{}.
\newblock
{\BBOQ}\APACrefatitle {The Hybrid Local Ensemble Transform Kalman Filter} {The hybrid local ensemble transform kalman filter}.{\BBCQ}
\newblock
\APACjournalVolNumPages{Monthly Weather Review}{142}{6}{2139 - 2149}.
\newblock
\begin{APACrefURL} \url{https://journals.ametsoc.org/view/journals/mwre/142/6/mwr-d-13-00131.1.xml} \end{APACrefURL}
\newblock
\begin{APACrefDOI} \doi{https://doi.org/10.1175/MWR-D-13-00131.1} \end{APACrefDOI}
\PrintBackRefs{\CurrentBib}

\bibitem [\protect \citeauthoryear {%
Penny%
}{%
Penny%
}{%
{\protect \APACyear {2017}}%
}]{%
penny2017}
\APACinsertmetastar {%
penny2017}%
\begin{APACrefauthors}%
Penny, S\BPBI G.%
\end{APACrefauthors}%
\unskip\
\newblock
\APACrefYearMonthDay{2017}{}{}.
\newblock
{\BBOQ}\APACrefatitle {{Mathematical foundations of hybrid data assimilation from a synchronization perspective}} {{Mathematical foundations of hybrid data assimilation from a synchronization perspective}}.{\BBCQ}
\newblock
\APACjournalVolNumPages{Chaos: An Interdisciplinary Journal of Nonlinear Science}{27}{12}{126801}.
\newblock
\begin{APACrefURL} \url{https://doi.org/10.1063/1.5001819} \end{APACrefURL}
\newblock
\begin{APACrefDOI} \doi{10.1063/1.5001819} \end{APACrefDOI}
\PrintBackRefs{\CurrentBib}

\bibitem [\protect \citeauthoryear {%
Penny%
\ \protect \BOthers {.}}{%
Penny%
\ \protect \BOthers {.}}{%
{\protect \APACyear {2022}}%
}]{%
penny_integrating_2022}
\APACinsertmetastar {%
penny_integrating_2022}%
\begin{APACrefauthors}%
Penny, S\BPBI G.%
, Smith, T\BPBI A.%
, Chen, T\BHBI C.%
, Platt, J\BPBI A.%
, Lin, H\BHBI Y.%
, Goodliff, M.%
\BCBL {}\ \BBA {} Abarbanel, H\BPBI D\BPBI I.%
\end{APACrefauthors}%
\unskip\
\newblock
\APACrefYearMonthDay{2022}{}{}.
\newblock
{\BBOQ}\APACrefatitle {Integrating {Recurrent} {Neural} {Networks} {With} {Data} {Assimilation} for {Scalable} {Data}-{Driven} {State} {Estimation}} {Integrating {Recurrent} {Neural} {Networks} {With} {Data} {Assimilation} for {Scalable} {Data}-{Driven} {State} {Estimation}}.{\BBCQ}
\newblock
\APACjournalVolNumPages{Journal of Advances in Modeling Earth Systems}{14}{3}{e2021MS002843}.
\newblock
\begin{APACrefURL} [{2023-10-12}]\url{https://onlinelibrary.wiley.com/doi/abs/10.1029/2021MS002843} \end{APACrefURL}
\newblock
\APACrefnote{\_eprint: https://onlinelibrary.wiley.com/doi/pdf/10.1029/2021MS002843}
\newblock
\begin{APACrefDOI} \doi{10.1029/2021MS002843} \end{APACrefDOI}
\PrintBackRefs{\CurrentBib}

\bibitem [\protect \citeauthoryear {%
Platt%
, Penny%
, Smith%
\BCBL {}\ \BBA {} Abarbanel%
}{%
Platt%
\ \protect \BOthers {.}}{%
{\protect \APACyear {2023}}%
}]{%
platt2023}
\APACinsertmetastar {%
platt2023}%
\begin{APACrefauthors}%
Platt, J\BPBI A.%
, Penny, S\BPBI G.%
, Smith, T\BPBI A.%
\BCBL {}\ \BBA {} Abarbanel, T\BHBI C\BPBI C\BPBI H\BPBI D\BPBI I.%
\end{APACrefauthors}%
\unskip\
\newblock
\APACrefYearMonthDay{2023}{}{}.
\newblock
{\BBOQ}\APACrefatitle {Constraining chaos: {Enforcing} dynamical invariants in the training of reservoir computers} {Constraining chaos: {Enforcing} dynamical invariants in the training of reservoir computers}.{\BBCQ}
\newblock
\APACjournalVolNumPages{Chaos: An Interdisciplinary Journal of Nonlinear Science}{33}{10}{103107}.
\newblock
\begin{APACrefURL} [{2023-10-11}]\url{https://doi.org/10.1063/5.0156999} \end{APACrefURL}
\newblock
\begin{APACrefDOI} \doi{10.1063/5.0156999} \end{APACrefDOI}
\PrintBackRefs{\CurrentBib}

\bibitem [\protect \citeauthoryear {%
Platt%
, Penny%
, Smith%
, Chen%
\BCBL {}\ \BBA {} Abarbanel%
}{%
Platt%
\ \protect \BOthers {.}}{%
{\protect \APACyear {2022}}%
}]{%
platt_systematic_2022}
\APACinsertmetastar {%
platt_systematic_2022}%
\begin{APACrefauthors}%
Platt, J\BPBI A.%
, Penny, S\BPBI G.%
, Smith, T\BPBI A.%
, Chen, T\BHBI C.%
\BCBL {}\ \BBA {} Abarbanel, H\BPBI D\BPBI I.%
\end{APACrefauthors}%
\unskip\
\newblock
\APACrefYearMonthDay{2022}{}{}.
\newblock
{\BBOQ}\APACrefatitle {A systematic exploration of reservoir computing for forecasting complex spatiotemporal dynamics} {A systematic exploration of reservoir computing for forecasting complex spatiotemporal dynamics}.{\BBCQ}
\newblock
\APACjournalVolNumPages{Neural Networks}{153}{}{530--552}.
\newblock
\begin{APACrefURL} [{2023-10-12}]\url{https://www.sciencedirect.com/science/article/pii/S0893608022002404} \end{APACrefURL}
\newblock
\begin{APACrefDOI} \doi{10.1016/j.neunet.2022.06.025} \end{APACrefDOI}
\PrintBackRefs{\CurrentBib}

\bibitem [\protect \citeauthoryear {%
Platt%
, Wong%
, Clark%
, Penny%
\BCBL {}\ \BBA {} Abarbanel%
}{%
Platt%
\ \protect \BOthers {.}}{%
{\protect \APACyear {2021}}%
}]{%
platt_robust_2021}
\APACinsertmetastar {%
platt_robust_2021}%
\begin{APACrefauthors}%
Platt, J\BPBI A.%
, Wong, A.%
, Clark, R.%
, Penny, S\BPBI G.%
\BCBL {}\ \BBA {} Abarbanel, H\BPBI D\BPBI I.%
\end{APACrefauthors}%
\unskip\
\newblock
\APACrefYearMonthDay{2021}{}{}.
\newblock
{\BBOQ}\APACrefatitle {Robust forecasting using predictive generalized synchronization in reservoir computing} {Robust forecasting using predictive generalized synchronization in reservoir computing}.{\BBCQ}
\newblock
\APACjournalVolNumPages{Chaos: An Interdisciplinary Journal of Nonlinear Science}{31}{12}{123118}.
\newblock
\begin{APACrefURL} [{2023-10-12}]\url{https://doi.org/10.1063/5.0066013} \end{APACrefURL}
\newblock
\begin{APACrefDOI} \doi{10.1063/5.0066013} \end{APACrefDOI}
\PrintBackRefs{\CurrentBib}

\bibitem [\protect \citeauthoryear {%
Rasp%
\ \protect \BOthers {.}}{%
Rasp%
\ \protect \BOthers {.}}{%
{\protect \APACyear {2024}}%
}]{%
rasp_weatherbench_2024}
\APACinsertmetastar {%
rasp_weatherbench_2024}%
\begin{APACrefauthors}%
Rasp, S.%
, Hoyer, S.%
, Merose, A.%
, Langmore, I.%
, Battaglia, P.%
, Russel, T.%
\BDBL {}Sha, F.%
\end{APACrefauthors}%
\unskip\
\newblock
\APACrefYearMonthDay{2024}{{\APACmonth{01}}}{}.
\newblock
\APACrefbtitle {{WeatherBench} 2: {A} benchmark for the next generation of data-driven global weather models.} {{WeatherBench} 2: {A} benchmark for the next generation of data-driven global weather models.}
\newblock
\APACaddressPublisher{}{arXiv}.
\newblock
\begin{APACrefURL} [{2024-04-05}]\url{http://arxiv.org/abs/2308.15560} \end{APACrefURL}
\newblock
\APACrefnote{arXiv:2308.15560 [physics]}
\newblock
\begin{APACrefDOI} \doi{10.48550/arXiv.2308.15560} \end{APACrefDOI}
\PrintBackRefs{\CurrentBib}

\bibitem [\protect \citeauthoryear {%
Reinhold%
\ \BBA {} Pierrehumbert%
}{%
Reinhold%
\ \BBA {} Pierrehumbert%
}{%
{\protect \APACyear {1982}}%
}]{%
reinhold_dynamics_1982}
\APACinsertmetastar {%
reinhold_dynamics_1982}%
\begin{APACrefauthors}%
Reinhold, B\BPBI B.%
\BCBT {}\ \BBA {} Pierrehumbert, R\BPBI T.%
\end{APACrefauthors}%
\unskip\
\newblock
\APACrefYearMonthDay{1982}{}{}.
\newblock
{\BBOQ}\APACrefatitle {Dynamics of {Weather} {Regimes}: {Quasi}-{Stationary} {Waves} and {Blocking}} {Dynamics of {Weather} {Regimes}: {Quasi}-{Stationary} {Waves} and {Blocking}}.{\BBCQ}
\newblock
\APACjournalVolNumPages{Monthly Weather Review}{110}{9}{1105--1145}.
\newblock
\begin{APACrefURL} [{2023-10-11}]\url{https://journals.ametsoc.org/view/journals/mwre/110/9/1520-0493_1982_110_1105_dowrqs_2_0_co_2.xml} \end{APACrefURL}
\newblock
\APACrefnote{Publisher: American Meteorological Society Section: Monthly Weather Review}
\newblock
\begin{APACrefDOI} \doi{10.1175/1520-0493(1982)110<1105:DOWRQS>2.0.CO;2} \end{APACrefDOI}
\PrintBackRefs{\CurrentBib}

\bibitem [\protect \citeauthoryear {%
Shampine%
}{%
Shampine%
}{%
{\protect \APACyear {1986}}%
}]{%
shampine1986some}
\APACinsertmetastar {%
shampine1986some}%
\begin{APACrefauthors}%
Shampine, L\BPBI F.%
\end{APACrefauthors}%
\unskip\
\newblock
\APACrefYearMonthDay{1986}{}{}.
\newblock
{\BBOQ}\APACrefatitle {Some practical Runge-Kutta formulas} {Some practical runge-kutta formulas}.{\BBCQ}
\newblock
\APACjournalVolNumPages{Mathematics of computation}{46}{173}{135--150}.
\PrintBackRefs{\CurrentBib}

\bibitem [\protect \citeauthoryear {%
Solvik%
\ \BBA {} Penny%
}{%
Solvik%
\ \BBA {} Penny%
}{%
{\protect \APACyear {2024}}%
{\protect \APACexlab {{\protect \BCnt {1}}}}}]{%
solvik_stevepnydataassimbench-examples_2024}
\APACinsertmetastar {%
solvik_stevepnydataassimbench-examples_2024}%
\begin{APACrefauthors}%
Solvik, K.%
\BCBT {}\ \BBA {} Penny, S.%
\end{APACrefauthors}%
\unskip\
\newblock
\APACrefYearMonthDay{2024{\protect \BCnt {1}}}{{\APACmonth{08}}}{}.
\newblock
\APACrefbtitle {{StevePny}/{DataAssimBench}-{Examples}: v0.1.1-alpha.} {{StevePny}/{DataAssimBench}-{Examples}: v0.1.1-alpha.}
\newblock
\APACaddressPublisher{}{Zenodo}.
\newblock
\begin{APACrefURL} [{2024-08-02}]\url{https://zenodo.org/records/13158546} \end{APACrefURL}
\newblock
\begin{APACrefDOI} \doi{10.5281/zenodo.13158546} \end{APACrefDOI}
\PrintBackRefs{\CurrentBib}

\bibitem [\protect \citeauthoryear {%
Solvik%
\ \BBA {} Penny%
}{%
Solvik%
\ \BBA {} Penny%
}{%
{\protect \APACyear {2024}}%
{\protect \APACexlab {{\protect \BCnt {2}}}}}]{%
solvik_stevepnydataassimbench_2024}
\APACinsertmetastar {%
solvik_stevepnydataassimbench_2024}%
\begin{APACrefauthors}%
Solvik, K.%
\BCBT {}\ \BBA {} Penny, S.%
\end{APACrefauthors}%
\unskip\
\newblock
\APACrefYearMonthDay{2024{\protect \BCnt {2}}}{{\APACmonth{08}}}{}.
\newblock
\APACrefbtitle {{StevePny}/{DataAssimBench}: v0.1.1-alpha.} {{StevePny}/{DataAssimBench}: v0.1.1-alpha.}
\newblock
\APACaddressPublisher{}{Zenodo}.
\newblock
\begin{APACrefURL} [{2024-08-02}]\url{https://zenodo.org/records/13158498} \end{APACrefURL}
\newblock
\begin{APACrefDOI} \doi{10.5281/zenodo.13158498} \end{APACrefDOI}
\PrintBackRefs{\CurrentBib}

\bibitem [\protect \citeauthoryear {%
van~der Vorst%
}{%
van~der Vorst%
}{%
{\protect \APACyear {1992}}%
}]{%
vanderVorst92}
\APACinsertmetastar {%
vanderVorst92}%
\begin{APACrefauthors}%
van~der Vorst, H\BPBI A.%
\end{APACrefauthors}%
\unskip\
\newblock
\APACrefYearMonthDay{1992}{}{}.
\newblock
{\BBOQ}\APACrefatitle {Bi-CGSTAB: A Fast and Smoothly Converging Variant of Bi-CG for the Solution of Nonsymmetric Linear Systems} {Bi-cgstab: A fast and smoothly converging variant of bi-cg for the solution of nonsymmetric linear systems}.{\BBCQ}
\newblock
\APACjournalVolNumPages{SIAM Journal on Scientific and Statistical Computing}{13}{2}{631-644}.
\newblock
\begin{APACrefURL} \url{https://doi.org/10.1137/0913035} \end{APACrefURL}
\newblock
\begin{APACrefDOI} \doi{10.1137/0913035} \end{APACrefDOI}
\PrintBackRefs{\CurrentBib}

\bibitem [\protect \citeauthoryear {%
Vlachas%
\ \protect \BOthers {.}}{%
Vlachas%
\ \protect \BOthers {.}}{%
{\protect \APACyear {2020}}%
}]{%
Vlachas20}
\APACinsertmetastar {%
Vlachas20}%
\begin{APACrefauthors}%
Vlachas, P.%
, Pathak, J.%
, Hunt, B.%
, Sapsis, T.%
, Girvan, M.%
, Ott, E.%
\BCBL {}\ \BBA {} Koumoutsakos, P.%
\end{APACrefauthors}%
\unskip\
\newblock
\APACrefYearMonthDay{2020}{}{}.
\newblock
{\BBOQ}\APACrefatitle {Backpropagation algorithms and Reservoir Computing in Recurrent Neural Networks for the forecasting of complex spatiotemporal dynamics} {Backpropagation algorithms and reservoir computing in recurrent neural networks for the forecasting of complex spatiotemporal dynamics}.{\BBCQ}
\newblock
\APACjournalVolNumPages{Neural Networks}{126}{}{191 - 217}.
\newblock
\begin{APACrefDOI} \doi{https://doi.org/10.1016/j.neunet.2020.02.016} \end{APACrefDOI}
\PrintBackRefs{\CurrentBib}

\bibitem [\protect \citeauthoryear {%
Wang%
, Droegemeier%
\BCBL {}\ \BBA {} White%
}{%
Wang%
\ \protect \BOthers {.}}{%
{\protect \APACyear {1998}}%
}]{%
wang_adjoint_1998}
\APACinsertmetastar {%
wang_adjoint_1998}%
\begin{APACrefauthors}%
Wang, Z.%
, Droegemeier, K.%
\BCBL {}\ \BBA {} White, L.%
\end{APACrefauthors}%
\unskip\
\newblock
\APACrefYearMonthDay{1998}{}{}.
\newblock
{\BBOQ}\APACrefatitle {The {Adjoint} {Newton} {Algorithm} for {Large}-{Scale} {Unconstrained} {Optimization} in {Meteorology} {Applications}} {The {Adjoint} {Newton} {Algorithm} for {Large}-{Scale} {Unconstrained} {Optimization} in {Meteorology} {Applications}}.{\BBCQ}
\newblock
\APACjournalVolNumPages{Computational Optimization and Applications}{10}{3}{283--320}.
\newblock
\begin{APACrefURL} [{2024-01-11}]\url{https://doi.org/10.1023/A:1018321307393} \end{APACrefURL}
\newblock
\begin{APACrefDOI} \doi{10.1023/A:1018321307393} \end{APACrefDOI}
\PrintBackRefs{\CurrentBib}

\bibitem [\protect \citeauthoryear {%
Wang%
, Navon%
, Zou%
\BCBL {}\ \BBA {} Le~Dimet%
}{%
Wang%
\ \protect \BOthers {.}}{%
{\protect \APACyear {1995}}%
}]{%
wang_truncated_1995}
\APACinsertmetastar {%
wang_truncated_1995}%
\begin{APACrefauthors}%
Wang, Z.%
, Navon, I\BPBI M.%
, Zou, X.%
\BCBL {}\ \BBA {} Le~Dimet, F\BPBI X.%
\end{APACrefauthors}%
\unskip\
\newblock
\APACrefYearMonthDay{1995}{}{}.
\newblock
{\BBOQ}\APACrefatitle {A truncated {Newton} optimization algorithm in meteorology applications with analytic {Hessian}/vector products} {A truncated {Newton} optimization algorithm in meteorology applications with analytic {Hessian}/vector products}.{\BBCQ}
\newblock
\APACjournalVolNumPages{Computational Optimization and Applications}{4}{3}{241--262}.
\newblock
\begin{APACrefURL} [{2024-01-11}]\url{http://link.springer.com/10.1007/BF01300873} \end{APACrefURL}
\newblock
\begin{APACrefDOI} \doi{10.1007/BF01300873} \end{APACrefDOI}
\PrintBackRefs{\CurrentBib}

\bibitem [\protect \citeauthoryear {%
Zou%
\ \protect \BOthers {.}}{%
Zou%
\ \protect \BOthers {.}}{%
{\protect \APACyear {1993}}%
}]{%
zou_numerical_1993}
\APACinsertmetastar {%
zou_numerical_1993}%
\begin{APACrefauthors}%
Zou, X.%
, Navon, I\BPBI M.%
, Berger, M.%
, Phua, K\BPBI H.%
, Schlick, T.%
\BCBL {}\ \BBA {} Le~Dimet, F\BPBI X.%
\end{APACrefauthors}%
\unskip\
\newblock
\APACrefYearMonthDay{1993}{}{}.
\newblock
{\BBOQ}\APACrefatitle {Numerical {Experience} with {Limited}-{Memory} {Quasi}-{Newton} and {Truncated} {Newton} {Methods}} {Numerical {Experience} with {Limited}-{Memory} {Quasi}-{Newton} and {Truncated} {Newton} {Methods}}.{\BBCQ}
\newblock
\APACjournalVolNumPages{SIAM Journal on Optimization}{3}{3}{582--608}.
\newblock
\begin{APACrefURL} [{2024-01-11}]\url{http://epubs.siam.org/doi/10.1137/0803029} \end{APACrefURL}
\newblock
\begin{APACrefDOI} \doi{10.1137/0803029} \end{APACrefDOI}
\PrintBackRefs{\CurrentBib}

\end{thebibliography}


\begin{thebibliography}{}

\bibitem [\protect \citeauthoryear {%
Bergstra%
, Komer%
, Eliasmith%
, Yamins%
\BCBL {}\ \BBA {} Cox%
}{%
Bergstra%
\ \protect \BOthers {.}}{%
{\protect \APACyear {2015}}%
}]{%
Bergstra_2015}
\APACinsertmetastar {%
Bergstra_2015}%
\begin{APACrefauthors}%
Bergstra, J.%
, Komer, B.%
, Eliasmith, C.%
, Yamins, D.%
\BCBL {}\ \BBA {} Cox, D\BPBI D.%
\end{APACrefauthors}%
\unskip\
\newblock
\APACrefYearMonthDay{2015}{July}{}.
\newblock
{\BBOQ}\APACrefatitle {Hyperopt: a Python library for model selection and hyperparameter optimization} {Hyperopt: a python library for model selection and hyperparameter optimization}.{\BBCQ}
\newblock
\APACjournalVolNumPages{Computational Science \& Discovery}{8}{1}{014008}.
\newblock
\begin{APACrefURL} \url{https://dx.doi.org/10.1088/1749-4699/8/1/014008} \end{APACrefURL}
\newblock
\begin{APACrefDOI} \doi{10.1088/1749-4699/8/1/014008} \end{APACrefDOI}
\PrintBackRefs{\CurrentBib}

\bibitem [\protect \citeauthoryear {%
Demaeyer%
, Cruz%
\BCBL {}\ \BBA {} Vannitsem%
}{%
Demaeyer%
\ \protect \BOthers {.}}{%
{\protect \APACyear {2020}}%
}]{%
demaeyer_qgs_2020}
\APACinsertmetastar {%
demaeyer_qgs_2020}%
\begin{APACrefauthors}%
Demaeyer, J.%
, Cruz, L\BPBI D.%
\BCBL {}\ \BBA {} Vannitsem, S.%
\end{APACrefauthors}%
\unskip\
\newblock
\APACrefYearMonthDay{2020}{{\APACmonth{12}}}{}.
\newblock
{\BBOQ}\APACrefatitle {qgs: {A} flexible {Python} framework of reduced-order multiscale climate models} {qgs: {A} flexible {Python} framework of reduced-order multiscale climate models}.{\BBCQ}
\newblock
\APACjournalVolNumPages{Journal of Open Source Software}{5}{56}{2597}.
\newblock
\begin{APACrefURL} [{2023-10-11}]\url{https://joss.theoj.org/papers/10.21105/joss.02597} \end{APACrefURL}
\newblock
\begin{APACrefDOI} \doi{10.21105/joss.02597} \end{APACrefDOI}
\PrintBackRefs{\CurrentBib}

\bibitem [\protect \citeauthoryear {%
Liaw%
\ \protect \BOthers {.}}{%
Liaw%
\ \protect \BOthers {.}}{%
{\protect \APACyear {2018}}%
}]{%
liaw2018tune}
\APACinsertmetastar {%
liaw2018tune}%
\begin{APACrefauthors}%
Liaw, R.%
, Liang, E.%
, Nishihara, R.%
, Moritz, P.%
, Gonzalez, J\BPBI E.%
\BCBL {}\ \BBA {} Stoica, I.%
\end{APACrefauthors}%
\unskip\
\newblock
\APACrefYearMonthDay{2018}{}{}.
\newblock
\APACrefbtitle {Tune: A Research Platform for Distributed Model Selection and Training.} {Tune: A research platform for distributed model selection and training.}
\PrintBackRefs{\CurrentBib}

\bibitem [\protect \citeauthoryear {%
Platt%
, Penny%
, Smith%
\BCBL {}\ \BBA {} Abarbanel%
}{%
Platt%
\ \protect \BOthers {.}}{%
{\protect \APACyear {2023}}%
}]{%
platt2023}
\APACinsertmetastar {%
platt2023}%
\begin{APACrefauthors}%
Platt, J\BPBI A.%
, Penny, S\BPBI G.%
, Smith, T\BPBI A.%
\BCBL {}\ \BBA {} Abarbanel, T\BHBI C\BPBI C\BPBI H\BPBI D\BPBI I.%
\end{APACrefauthors}%
\unskip\
\newblock
\APACrefYearMonthDay{2023}{}{}.
\newblock
{\BBOQ}\APACrefatitle {Constraining chaos: {Enforcing} dynamical invariants in the training of reservoir computers} {Constraining chaos: {Enforcing} dynamical invariants in the training of reservoir computers}.{\BBCQ}
\newblock
\APACjournalVolNumPages{Chaos: An Interdisciplinary Journal of Nonlinear Science}{33}{10}{103107}.
\newblock
\begin{APACrefURL} [{2023-10-11}]\url{https://doi.org/10.1063/5.0156999} \end{APACrefURL}
\newblock
\begin{APACrefDOI} \doi{10.1063/5.0156999} \end{APACrefDOI}
\PrintBackRefs{\CurrentBib}

\end{thebibliography}
%




%
%
%
%
%

\end{document}


%
%


\title{Supporting Information for ``4D-Var using Hessian approximation and backpropagation applied to automatically-differentiable numerical and machine learning model''}
%
%

%
%

\authors{Kylen Solvik\affil{1}, Stephen G. Penny\affil{2}\affil{3}, Stephan Hoyer\affil{4}}

\affiliation{1}{University of Colorado Boulder, USA}
\affiliation{2}{Sofar Ocean, San Francisco CA, USA}
\affiliation{3}{Cooperative Institute for Research in Environmental Sciences at the University of Colorado Boulder}
\affiliation{4}{Google Research, Mountain View, CA 94043}

%
%

%

\begin{article}

%
%

\noindent\textbf{Contents of this file}
\begin{enumerate}
\item Table S1

\end{enumerate}
\noindent\textbf{Additional Supporting Information (Files uploaded separately)}
\begin{enumerate}
\item Captions for Datasets S1 to S2
\end{enumerate}

\noindent\textbf{Introduction}
Table S1 provides the pre-trained macro-parameters for the reservoir computing model used as a surrogate model for the quasi-geostrophic system experiments using the qgs Python package \cite{demaeyer_qgs_2020}. Dataset S1 contains the pre-LaTeX mark Error: Infinite shrinkage found in 'page'.trained micro-parameters (or model weights) for the reservoir computing model from \citeA{platt2023}. Dataset S2 contains the tuned learning rate and learning rate decay (``lr\_decay'') values used for the Backprop-4DVar data assimilation experiments, found by minimizing the RMSE on a validation set using Bayesian optimization via hyperopt \cite{Bergstra_2015} and RayTune \cite{liaw2018tune}. 
%


\noindent\textbf{Data Set S1.} Pre-trained reservoir computing micro-scale parameters (or model weights) stored as a pickle file (.pkl). The model is trained on a training dataset consisting of 100,000 time steps of the qgs model integration.


\noindent\textbf{Data Set S2.} Learning rate ($\alpha$, ``learning\_rate'') and learning rate decay (``lr\_decay'') values used for all Backprop-4DVar data assimilation experiments, determined using Bayesian optimization to minimize RMSE. 

%
%


%
%
%
%
%

\bibliography{si}

%
%
%
%
%

%
%

\end{article}

\clearpage



\begin{table}
\settablenum{S1} 
\caption{Macro-parameters for reservoir computing model from \citeA{platt2023}}
\centering
\begin{tabular}{l c}
\hline
 Macro-Parameter & Value\\
 \hline

 $Spectral Radius$  & 0.376752115791648\\
 $\alpha$ & 0.5343730100231164\\
 $\sigma_u$ & 0.9876577724115441\\
 $\sigma_b$ & 0.675882947305197\\
 $\beta$ & 1.65505395540213e-9\\
 $N$ & 2000\\
 $\rho_s$ & 0.01\\
\hline
\end{tabular}
\end{table}